\documentclass[letterpaper]{article} 
\usepackage[preprint]{aaai2027} 
\usepackage[hyphens]{url} 
\usepackage{graphicx} 
\usepackage{natbib} 
\usepackage{caption} 
\usepackage{amsmath,amssymb}
\usepackage{booktabs}
\usepackage{xr}
\RequirePackage{multicol}
\makeatletter
\newcommand{\appendixtableofcontents}{%
  \section*{Appendix Contents}%
  \setcounter{tocdepth}{2}%
  \begin{multicols}{2}%
    \@starttoc{atoc}%
  \end{multicols}%
}

\newcommand{\enableappendixtoc}{%
  \let\appendix@originalsection\section
  \let\appendix@originalsubsection\subsection
  \renewcommand{\section}[1]{%
    \appendix@originalsection{##1}%
    \addtocontents{atoc}{%
      \protect\contentsline{section}%
      {\protect\numberline{\thesection}##1}{\thepage}{}%
    }%
  }%
  \renewcommand{\subsection}[1]{%
    \appendix@originalsubsection{##1}%
    \addtocontents{atoc}{%
      \protect\contentsline{subsection}%
      {\protect\numberline{\thesubsection}##1}{\thepage}{}%
    }%
  }%
}

\newcommand{\disableappendixtoc}{%
  \let\section\appendix@originalsection
  \let\subsection\appendix@originalsubsection
}

\newcommand{\appendixtoccolumnbreak}{%
  \addtocontents{atoc}{\protect\columnbreak}%
}
\makeatother

\newif\ifwithappendix
\withappendixtrue
\ifwithappendix
  \newcommand{\suppref}[1]{\ref{#1}}
  \newcommand{\suppeqref}[1]{\eqref{#1}}
\else
  \makeatletter
  \let\saved@bibcite\bibcite
  \let\bibcite\@gobbletwo
  \let\bibcite\saved@bibcite
  \makeatother
  \newcommand{\suppref}[1]{\ref{supp-#1}}
  \newcommand{\suppeqref}[1]{\eqref{supp-#1}}
\fi

\providecommand{\method}{Cova-PINN}

\title{Cova-PINN: Cross-Domain Conservation Physics-Informed Neural Network for Fluid-Solid Conjugate Heat Transfer in Complex Geometries}
\author{Weizheng Zhang\equalcontrib\textsuperscript{1}, Xunjie Xie\equalcontrib\textsuperscript{1}, Hao Pan\textsuperscript{2}, and Lin Lu\textsuperscript{1}}
\affiliations{\textsuperscript{1}Shandong University\quad \textsuperscript{2}Tsinghua University}

\begin{document}

\maketitle

\begin{abstract}
Multi-domain physics-informed neural networks (PINNs) flexibly model medium-specific representations to solve fluid--solid conjugate heat transfer (CHT).
However, standard multi-domain PINNs enforce governing equations and interface conditions on separately sampled domain supports, which can yield plausible temperature fields but inaccurate end-to-end energy transfer and outlet temperatures.
We propose Cova-PINN, a multi-domain PINN framework that aligns conservation support with thermal interaction paths in complex geometries.
Cova-PINN jointly optimizes cross-domain composite control-volume balances at the local scale and paired-wall closure at the global exchanger scale.
We evaluate Cova-PINN on four triply periodic minimal surface (TPMS) heat exchangers and a geometrically distinct DualMS design against CHT-specific, optimization-oriented, and complex-geometry PINN baselines under a common protocol.
Relative to the closest baseline, MUSA-PINN-CHT, Cova-PINN reduces average outlet-temperature and device-level closure errors across the four TPMS topologies by $37.7\%$ and $60.2\%$, respectively, while also improving full-field and heat-duty accuracy, with consistent gains on DualMS.

\end{abstract}

\section{Introduction}

Physics-informed machine learning turns governing equations and boundary conditions into objectives for learning PDE-governed fields with limited high-fidelity supervision \cite{raissi2019physics,karniadakis2021physics}. PINNs are a widely used realization, and multiphysics variants often assign separate subnetworks to different media or state variables \cite{jagtap2020cpinn,jagtap2020xpinn,lee2025extended}. This modularity supports medium-specific representations, but local losses on separate spatial supports need not make the learned fields globally consistent---especially when a conserved quantity crosses several learned domains.

Conjugate heat transfer (CHT) makes this issue precisely diagnosable: at steady state without internal heat generation, heat released by the hot stream must cross the solid and be equal to that absorbed by the cold stream. Exact fields inherit this device-scale balance from the local equations and interface conditions, whereas PINNs optimize finite, nonconvex objectives with documented training pathologies \cite{krishnapriyan2021failure,wang2022ntk,hao2024pinnacle}. Because these objectives do not directly constrain outlet bulk temperatures or paired heat rates, plausible fields may still misrepresent exchanger-level transfer. We call this separation the \emph{cross-domain conservation gap}. It is especially consequential in complex exchangers with tortuous channels and distributed fluid--solid interaction paths.

Prior work improves domain decomposition and weak enforcement \cite{jagtap2020cpinn,jagtap2020xpinn,kharazmi2021hpvpinn}, multi-network CHT training \cite{lee2025extended}, pointwise optimization \cite{wu2024ropinn,duan2025copinn}, and conservation in complex internal flows \cite{zhang2026musa}. However, these methods do not explicitly align thermal conservation support with the complete hot-fluid--solid--cold-fluid path. The key design question is therefore not only whether to impose an integral constraint, but where its support lies relative to the interaction topology.

\begin{figure*}[t]
    \centering
    \includegraphics[width=0.99\textwidth,trim=0 105 0 0,clip]{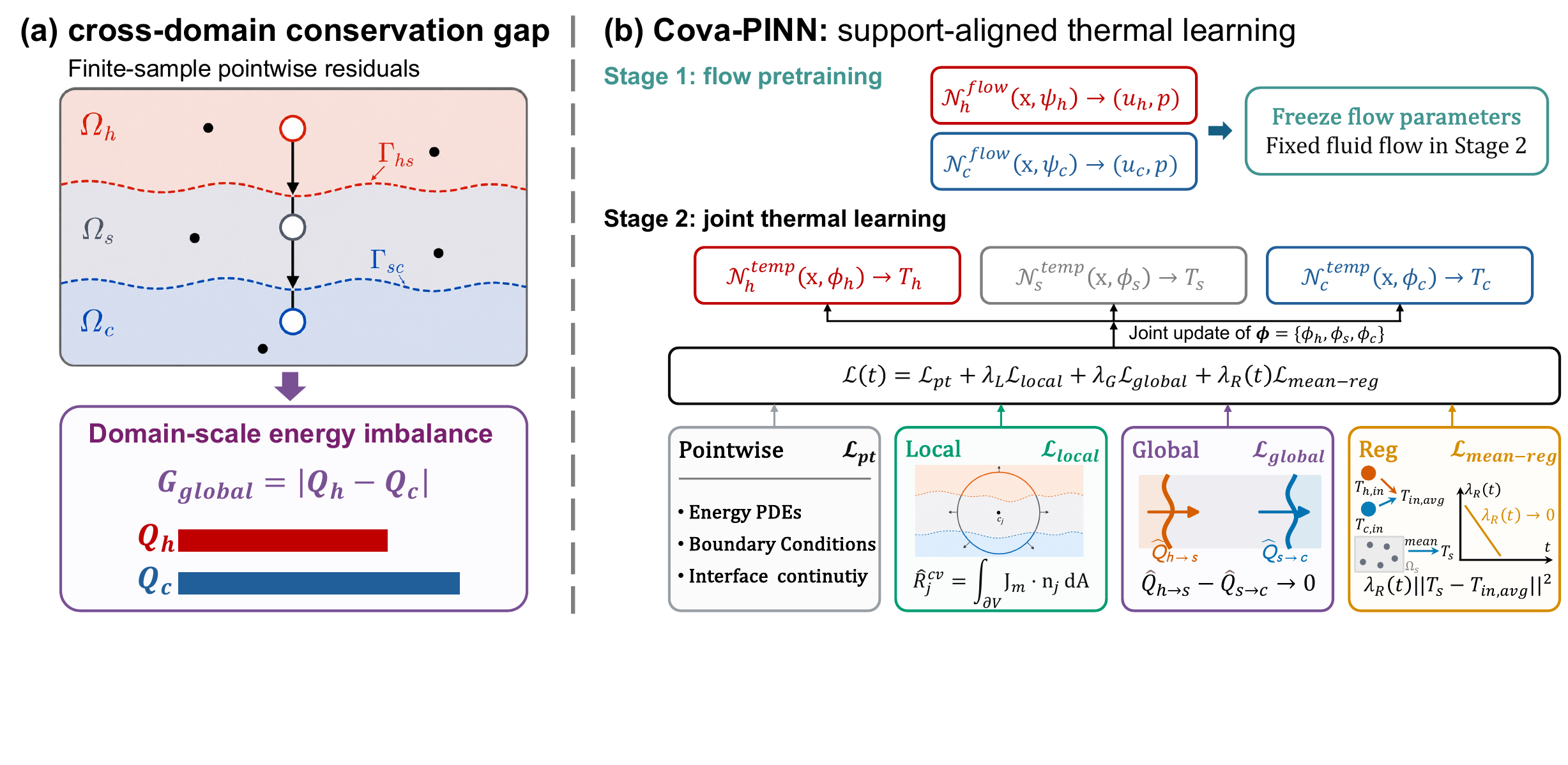}
    \caption{Overview of Cova-PINN. (a) Energy follows the hot-fluid--solid--cold-fluid path, while separately supported pointwise residuals need not close its device-scale balance. (b) After independently training and freezing the two MUSA-PINN flow networks, the three temperature networks are jointly optimized by pointwise CHT physics and coordinated local--global conservation objectives, with annealed solid-mean regularization during early training.}
    \label{fig:pipeline}
\end{figure*}

We introduce \textbf{Cova-PINN}, a five-network multi-domain framework whose conservation support follows this interaction path. After independently training and freezing the two MUSA-PINN flow networks \cite{zhang2026musa}, Cova-PINN jointly optimizes local, wall-centered composite-control-volume balances and global paired-wall closure between the solid-side heat rates. A solid-mean temperature regularizer conditions early optimization and then decays to zero. Figure~\ref{fig:pipeline} summarizes the framework.

Our contributions are threefold:
\begin{itemize}
    \item We demonstrate that domain-wise pointwise training can yield inaccurate exchanger-level energy transfer even when it reconstructs plausible temperature fields.
    \item We introduce a support-aligned global--local conservation co-optimization framework: cross-domain control volumes couple finite-region transfer, paired-wall closure coordinates exchanger-scale transfer, and annealed solid-mean regularization conditions early optimization.
    \item Across triply periodic minimal surface (TPMS) topologies and DualMS \cite{yeranee2022tpms,zhang2025dualms}, Cova-PINN reduces field, outlet, heat-duty, and closure errors; ablations support all three components.
\end{itemize}

\section{Related Work}

\paragraph{Multi-domain PINNs for conjugate heat transfer.}
Multi-network physics-informed models arise naturally when equations or material properties differ across subdomains. cPINN, XPINN, and FBPINN distribute a PDE solution across networks coupled through interfaces or overlapping supports \cite{jagtap2020cpinn,jagtap2020xpinn,moseley2023fbpinn}. More recently, MDPINN-GD partitions complex three-dimensional intersecting channels into regular subdomains and supplements their coupling with global inlet--outlet mass-flow conservation \cite{hwang2025mdpinn}. These methods demonstrate the complementary value of local decomposition and global constraints within a single physical domain. Multi-medium thermal models instead associate networks with material regions: M-PINN treats heterogeneous thermal layers \cite{zhang2022multidomain}, while I-PINNs and IG-PINNs target sharp interfaces and discontinuous coefficients \cite{sarma2024interface,zheng2026igpinn}. For CHT, modular PINNs have been applied to finned heat sinks and manifold microchannels \cite{lu2024finned,zhang2024microchannel}, and E-MPINN combines medium-specific subnetworks with physics-guided training \cite{lee2025extended}. We therefore select E-MPINN as the task-specific multi-network CHT baseline. Cova-PINN further aligns local conservation support with the complete fluid--solid--fluid interaction path and imposes paired-wall closure at the exchanger scale.

\paragraph{Weak-form and conservation-aware physics-informed solvers.}
A second line of research reformulates physics supervision over finite regions rather than enforcing differential residuals only at individual points. The Deep Ritz method minimizes variational energy functionals \cite{e2018deep}, weak adversarial networks enforce weak formulations using learned test functions \cite{zang2020weak}, and hp-VPINNs impose piecewise variational residuals over decomposed subdomains \cite{kharazmi2019variational,kharazmi2021hpvpinn}. More closely related to our setting, control-volume and finite-volume PINNs integrate governing equations over local cells and enforce the resulting flux balances on their boundaries, thereby strengthening finite-region conservation and reducing reliance on high-order pointwise derivatives \cite{wei2025ffvpinn,wei2025cvpinn}. These methods establish integral residuals and control-volume constraints as effective forms of physics supervision. A less explored question in multi-network multiphysics learning is how the support of such constraints should be chosen relative to the topology of physical interaction. Existing formulations typically apply weak constraints within a single PDE domain or according to a predefined geometric decomposition. In contrast, Cova-PINN places wall-centered composite supports across adjoining material regions along the hot-fluid--solid--cold-fluid interaction path, combines the intersected media in a shared local energy balance, and complements these local constraints with device-scale heat-rate closure across the paired heat-transfer walls.

\paragraph{PINN optimization in complex geometries.}
Even correctly specified continuous physics may be poorly realized by finite, nonconvex PINN training because of stiff gradient competition, uneven residual learning, and sampling-dependent failure modes \cite{wang2021gradient,krishnapriyan2021failure,wang2022ntk,hao2024pinnacle}. RoPINN expands pointwise optimization to calibrated neighborhoods, whereas CoPINN orders samples from easy to hard using a learned difficulty measure \cite{wu2024ropinn,duan2025copinn}; we include them as representative recent PINN optimization baselines. These methods improve how local objectives are sampled or optimized, but do not change which physical domains a conservation law couples. For tortuous internal flows, MUSA-PINN places multiscale weak-form control volumes along complex transport pathways \cite{zhang2026musa}. We use it as the complex-geometry flow baseline and closest structural comparator, while Cova-PINN extends conservation supervision from separate fluid domains to the coupled thermal interaction path. Taken together, prior work has advanced heterogeneous representation, interface transmission, integral conservation, and trainability in complex domains. Our work addresses their intersection by aligning conservation support with a multi-domain thermal interaction path.

\section{Problem Formulation}
\label{sec:problem}

\subsection{Physical Setting and Energy Transfer}

Figure~\ref{fig:pipeline}(a) shows the physical setting and introduces the notation used below. The heat exchanger contains a hot-fluid region $\Omega_h$, a separating solid $\Omega_s$, and a cold-fluid region $\Omega_c$. Their two contact surfaces are denoted by $\Gamma_{hs}$ (hot fluid--solid) and $\Gamma_{sc}$ (solid--cold fluid). These labels are needed because the three regions are represented by different temperature networks, while heat must pass consistently across both interfaces.

The same figure also shows the resulting energy-transfer path: heat leaves $\Omega_h$, crosses $\Gamma_{hs}$, conducts through $\Omega_s$, crosses $\Gamma_{sc}$, and enters $\Omega_c$. Thus, one physical process spans three separately learned domains. We consider steady forced convection with constant material properties, negligible buoyancy, and no internal heat generation; both streams use the same fluid.

We use dimensionless variables to keep the formulation compact. Temperature is scaled so that the hot and cold inlets have values $1$ and $0$, respectively. For a fluid stream $f\in\{h,c\}$, the P\'eclet number $Pe_f$ measures advection relative to thermal diffusion, while $\kappa_s$ is the solid-to-fluid conductivity ratio. The complete nondimensionalization, material parameters, and boundary conditions are given in Appendix~\suppref{app:problem}.

With no internal heat generation, the dimensionless energy equations are
\begin{equation}
\mathbf u_f\!\cdot\nabla T_f-\frac{1}{Pe_f}\nabla^2T_f=0,
\quad f\in\{h,c\},
\qquad
\nabla^2T_s=0.
\label{eq:dimensionless-energy}
\end{equation}
The first equation describes convection and diffusion in each fluid, and the second describes conduction in the solid. Because our method enforces energy balance over finite regions, we next require a quantity whose boundary integral measures the net energy crossing such a region. We therefore define the dimensionless energy-flux density
\begin{equation}
\begin{aligned}
\mathbf J_f&=Pe_f\mathbf u_fT_f-\nabla T_f,
&&f\in\{h,c\},\\
\mathbf J_s&=-\kappa_s\nabla T_s,\qquad
&&\nabla\!\cdot\mathbf J_m=0,\quad m\in\{h,s,c\}.
\end{aligned}
\label{eq:common-flux}
\end{equation}
Here $\mathbf J_f$ combines convective and conductive transport in a fluid, whereas $\mathbf J_s$ contains only solid conduction. All three fluxes use a common scale, so their surface integrals can be added directly in the cross-domain control volumes introduced in Sec.~\ref{sec:method}.

Let $\mathbf n_f$ and $\mathbf n_s$ point outward from the adjoining fluid and solid regions. Conjugate heat transfer requires
\begin{equation}
\begin{aligned}
\mathbf u_f&=\mathbf 0,\qquad T_f=T_s,\\
\mathbf J_f\!\cdot\mathbf n_f
+\mathbf J_s\!\cdot\mathbf n_s&=0
\qquad\text{on }\Gamma_{hs}\text{ and }\Gamma_{sc},
\end{aligned}
\label{eq:dimensionless-interface}
\end{equation}
These three conditions impose no slip, temperature continuity, and normal heat-flux continuity, respectively. Conditions on the remaining boundaries are tabulated in Appendix~\suppref{app:problem}.

\subsection{Sequential Multi-Network Learning Setting}
\label{sec:sequential-setting}

We represent the two flow fields and three temperature fields with independent neural networks:
\begin{equation}
\begin{aligned}
(\mathbf u_f,p_f)&=\mathcal N_f^{\mathrm{flow}}(\mathbf x;\psi_f),
&f&\in\{h,c\},\\
T_m&=\mathcal N_m^{\mathrm{temp}}(\mathbf x;\phi_m),
&m&\in\{h,s,c\}.
\end{aligned}
\label{eq:five-networks}
\end{equation}
Let $\boldsymbol{\phi}:=(\phi_h,\phi_s,\phi_c)$ collect the parameters of the three temperature networks.
The flow networks are trained first, after which their velocity predictions $\bar{\mathbf u}_h$ and $\bar{\mathbf u}_c$ are frozen. The thermal stage therefore solves a sequential CHT problem with fixed advection rather than a bidirectionally coupled thermofluid problem. The flow-stage PDE, boundary, and complete MUSA objectives are defined in Appendix~\suppref{app:configuration}, Eqs.~\suppeqref{eq:sup-flow-pde-loss}--\suppeqref{eq:sup-musa-flow-loss}.

The standard pointwise thermal objective penalizes Eq.~\eqref{eq:dimensionless-energy}, the thermal boundary conditions, and the two thermal interface conditions in Eq.~\eqref{eq:dimensionless-interface}:
\begin{equation}
\mathcal L_{\mathrm{pt}}=
\lambda_{\mathrm{PDE}}\mathcal L_{\mathrm{PDE}}^T
+\lambda_{\mathrm{bc}}\mathcal L_{\mathrm{bc}}^T
+\lambda_{\Gamma,T}\mathcal L_{\Gamma,T}
+\lambda_{\Gamma,q}\mathcal L_{\Gamma,q}.
\label{eq:pointwise-loss}
\end{equation}
Here $\mathcal L_{\mathrm{PDE}}^T$, $\mathcal L_{\mathrm{bc}}^T$, $\mathcal L_{\Gamma,T}$, and $\mathcal L_{\Gamma,q}$ are empirical mean-squared residuals for the energy equations, thermal boundary conditions, interface temperature continuity, and interface normal-flux continuity, respectively. Each domain or boundary component is averaged over its own sample set before weighting. Their complete samplewise definitions are given in Appendix~\suppref{app:configuration}, Eqs.~\suppeqref{eq:sup-thermal-residuals}--\suppeqref{eq:sup-pointwise-total}; network and optimization settings and loss weights are listed in Tables~\suppref{tab:sup-training} and \suppref{tab:sup-loss-weights}. This common pointwise thermal objective underlies every CHT method in our comparison.

Because these residuals are evaluated on separate domain- and interface-specific sample sets, no individual pointwise residual directly spans the complete transfer path in Fig.~\ref{fig:pipeline}(a).

\subsection{End-to-End Conservation Requirement}

For the steady exchanger considered here, whose exterior solid boundary is adiabatic and whose domains contain no internal heat generation, the heat released by the hot stream must equal the heat absorbed by the cold stream. Let $Q_h$ and $Q_c$ denote these heat rates, inferred from the inlet and outlet bulk temperatures and normalized by $Q_0=C_{\min}\Delta T$, where $C_{\min}=\min(\dot m_hc_{p,f},\dot m_cc_{p,f})$. The device-level balance is
\begin{equation}
Q_h=Q_c,
\qquad
G_{\mathrm{global}}=|Q_h-Q_c|=0.
\label{eq:global-gap}
\end{equation}
We use three related but distinct levels of energy notation. The flux density $\mathbf J_m$ enters finite-region control-volume balances. The solid-side wall integrals $\widehat Q_{h\to s}$ and $\widehat Q_{s\to c}$, normalized by $Q_{\mathrm{ref}}$, form the Global training objective in Sec.~\ref{sec:method}. By contrast, $Q_h$ and $Q_c$ are inlet--outlet heat rates normalized by $Q_0$ and are used only for device-level evaluation. Their complete bulk-temperature definitions are given in Appendix~\suppref{app:metrics}.

Integrating the local energy equations over the three media and applying interface flux continuity yields Eq.~\eqref{eq:global-gap}. The finite-sample multi-network objective in Eq.~\eqref{eq:pointwise-loss}, however, evaluates domain and interface residuals on separate spatial supports and does not explicitly optimize this cross-domain integral relation. Therefore, minimizing the pointwise objective does not by itself guarantee device-level energy conservation, i.e.:
\begin{equation}
\mathcal L_{\mathrm{pt}}(\boldsymbol{\phi}) \rightarrow 0
\quad\not\Rightarrow\quad
G_{\mathrm{global}}(\boldsymbol{\phi})\rightarrow 0.
\label{eq:pointwise-global-gap}
\end{equation}
This separation can manifest as inaccurate outlet bulk temperatures and nonzero $G_{\mathrm{global}}$. We call it the \emph{cross-domain conservation gap}. Section~\ref{sec:method} turns the corresponding finite-region and device-level conservation relations into optimization objectives.

\section{Cova-PINN}
\label{sec:method}

\subsection{Overview and Global--Local Co-Optimization}

Figure~\ref{fig:pipeline}(b) summarizes the two-stage procedure. Under the temperature-independent flow assumption in Sec.~\ref{sec:sequential-setting}, Stage~1 trains the two flow networks with MUSA-PINN \cite{zhang2026musa} and freezes $\bar{\mathbf u}_h$ and $\bar{\mathbf u}_c$. Stage~2 updates only the thermal parameters $\boldsymbol{\phi}$ and retains the standard energy equations, boundary conditions, and interface transmission in the \emph{Pointwise} block. On this basis, Cova-PINN performs global--local conservation co-optimization: \emph{Local} couples the temperature networks through shared finite-region balances along the thermal interaction path, while \emph{Global} closes the accumulated heat transfer at the exchanger scale. \emph{Reg} supports this two-scale objective by anchoring the solid-domain mean temperature against early drift toward implausible levels; its weight then vanishes so that the final solution is governed by the CHT and conservation objectives.

\subsection{Local Cross-Domain Composite Control Volumes}

The \emph{Local} module addresses a limitation of $\mathcal L_{\mathrm{pt}}$: its domain equations and pairwise interface conditions are evaluated on separate supports and therefore provide no shared finite-region balance over the interacting media. We construct such supports by sampling $N_{\mathrm{cv}}$ spherical parent regions $B_j=B(\mathbf c_j,r)$, $j=1,\ldots,N_{\mathrm{cv}}$, with centers $\mathbf c_j$ on the heat-transfer walls $\Gamma_{\mathrm w}:=\Gamma_{hs}\cup\Gamma_{sc}$. The radius is selected relative to the nominal wall thickness so that these wall-centered supports extend into adjoining material regions along the thermal interaction path. Intersections define $V_{m,j}=B_j\cap\Omega_m$ and the composite volume $V_j=B_j\cap\Omega$, whose outer boundary is partitioned into material-labeled patches $S_{m,j}=\partial V_j\cap\overline{\Omega}_m$. Spherical normals are analytic, enabling direct area-uniform flux quadrature; the implemented radius, boundary handling, and sampling details appear in Appendix~\suppref{app:local-cv}.

Using the common-scale energy fluxes in Eq.~\eqref{eq:common-flux}, integration over the material portions gives the compact balance
\begin{equation}
\begin{aligned}
0
&=\sum_{m\in\{h,s,c\}}
\int_{V_{m,j}}\nabla\!\cdot\mathbf J_m\,\mathrm dV\\
&\overset{\mathrm{Gauss}}{=}
\sum_{m\in\{h,s,c\}}
\int_{\partial V_{m,j}}
\mathbf J_m\!\cdot\mathbf n_{m,j}\,\mathrm dA\\
&\overset{\text{interface conservation}}{=}
\sum_{m\in\{h,s,c\}}
\int_{S_{m,j}}\mathbf J_m\!\cdot\mathbf n_j\,\mathrm dA
\;=:R_j^{\mathrm{cv}}(\boldsymbol{\phi}).
\end{aligned}
\label{eq:local_cv}
\end{equation}
No penetration removes advective interface transfer, while normal-flux continuity cancels internal interface contributions. Thus $\mathcal L_{\mathrm{local}}$ uses only the material-labeled outer patches; pointwise losses retain temperature and flux transmission.

Each surface integral is estimated by an area-uniform sample mean times the component area, without local-area weights. With $\widehat R_j^{\mathrm{cv}}$ denoting the fixed-scale estimator,
\begin{equation}
\mathcal L_{\mathrm{local}}
=\frac{1}{N_{\mathrm{cv}}}\sum_{j=1}^{N_{\mathrm{cv}}}
\left(\widehat R_j^{\mathrm{cv}}\right)^2.
\label{eq:local-loss}
\end{equation}
Each composite residual aggregates flux contributions from all material regions intersected by its support and jointly updates their corresponding temperature networks. Across the wall-centered support set, these coupled local updates connect the hot-fluid--solid--cold-fluid interaction path. Appendix~\suppref{app:local-cv} gives the complete estimator.

\subsection{Global Paired-Wall Heat-Rate Closure}

The \emph{Local} module coordinates cross-domain transfer over finite neighborhoods, but these local balances do not directly close the total heat entering and leaving the solid over the complete exchanger. The \emph{Global} module supplies this complementary device-scale constraint by pairing $\Gamma_{hs}$ and $\Gamma_{sc}$ and evaluating both heat rates from the solid. Using $\mathbf J_s=-\kappa_s\nabla T_s$ from Eq.~\eqref{eq:common-flux}, with $\mathbf n_s^h$ and $\mathbf n_s^c$ pointing from the solid toward the hot and cold fluids, the positive entering and leaving rates are
\begin{align}
\widehat Q_{h\to s}(\phi_s)
&=-\int_{\Gamma_{hs}}
\mathbf J_s\!\cdot\!\mathbf n_s^h\,\mathrm dA,\\
\widehat Q_{s\to c}(\phi_s)
&=\int_{\Gamma_{sc}}
\mathbf J_s\!\cdot\!\mathbf n_s^c\,\mathrm dA.
\end{align}
Steady energy balance requires these rates to match. Evaluating both from solid-side fluxes provides one accounting domain across the paired walls.

The corresponding residual and loss are
\begin{align}
R^{\mathrm{wall}}(\phi_s)
&=\widehat Q_{h\to s}-\widehat Q_{s\to c},\\
\mathcal L_{\mathrm{global}}
&=\left(R^{\mathrm{wall}}(\phi_s)\right)^2.
\label{eq:global-loss}
\end{align}
The loss directly updates $\phi_s$; pointwise normal-flux conditions in Eq.~\eqref{eq:dimensionless-interface} transmit its effect to the fluid temperature networks. The inlet--outlet gap $G_{\mathrm{global}}$, reported as $E_{\mathrm{clo}}$, remains an independent diagnostic rather than a training term. Estimator details are given in Appendix~\suppref{app:global-closure}.

\subsection{Annealed Solid-Mean Temperature Regularization}

The \emph{Local} and \emph{Global} modules form the core two-scale conservation objective. During early optimization, however, the independently initialized temperature networks can drift toward incompatible or implausible overall levels before cross-domain coupling is established. The \emph{Reg} module anchors the solid-domain mean temperature during this initial phase without prescribing its spatial structure:
\begin{equation}
\begin{aligned}
\overline T_s(\phi_s)
&=\frac{1}{|\Omega_s|}
\int_{\Omega_s}T_s(\mathbf x;\phi_s)\,\mathrm dV,\\
T_{\mathrm{in,avg}}
&=\frac{\dot m_hT_{h,\mathrm{in}}+\dot m_cT_{c,\mathrm{in}}}
{\dot m_h+\dot m_c}.
\end{aligned}
\label{eq:solid-mean-reference}
\end{equation}
Because both streams use the same fluid, this is also heat-capacity-rate weighting. For volume-uniform solid samples $\mathcal X_s^R$, $N_s^R=|\mathcal X_s^R|$,
\begin{equation}
\begin{aligned}
\widehat{\overline T}_s
&=\frac{1}{N_s^R}\sum_{\mathbf x_i\in\mathcal X_s^R}
T_s(\mathbf x_i;\phi_s),\\
\mathcal L_{\mathrm{mean\text{-}reg}}
&=\left(\widehat{\overline T}_s-T_{\mathrm{in,avg}}\right)^2.
\end{aligned}
\label{eq:solid-mean-regularizer}
\end{equation}
This zero-order regularizer preserves the spatial degrees of freedom of $T_s$ and influences the fluid networks only through the coupled physics losses.

At thermal update $t$, the complete objective is
\begin{equation}
\mathcal L(t)=\mathcal L_{\mathrm{pt}}
+\lambda_L\mathcal L_{\mathrm{local}}
+\lambda_G\mathcal L_{\mathrm{global}}
+\lambda_R(t)\mathcal L_{\mathrm{mean\text{-}reg}}.
\label{eq:total-loss}
\end{equation}
The conservation weights $\lambda_L$ and $\lambda_G$ remain fixed throughout thermal training. With $N_T$ thermal updates, $N_R=N_T/2$, and $\lambda_R^{(0)}=1$, $\lambda_R(t)=\lambda_R^{(0)}\max(1-t/N_R,0)$ decreases linearly to zero over the first half of training and remains inactive thereafter. The implemented loss weights and method-specific conservation quadrature are reported in Appendix~\suppref{app:configuration}.

\begin{table*}[t]
\centering

\scriptsize
\setlength{\tabcolsep}{2.2pt}
\renewcommand{\arraystretch}{1.04}
\resizebox{\textwidth}{!}{%
\begin{tabular}{lc*{4}{ccc}}
\toprule
& & \multicolumn{3}{c}{\textbf{Primitive (P)}} & \multicolumn{3}{c}{\textbf{Gyroid (G)}} & \multicolumn{3}{c}{\textbf{Diamond (D)}} & \multicolumn{3}{c}{\textbf{IWP}} \\
\cmidrule(lr){3-5}\cmidrule(lr){6-8}\cmidrule(lr){9-11}\cmidrule(lr){12-14}
\textbf{Method} & \textbf{Ref.}
& $E_{\mathrm{field}}$ (K) & $E_{\mathrm{out}}$ (K) & $E_{\mathrm{clo}}$ (\%)
& $E_{\mathrm{field}}$ (K) & $E_{\mathrm{out}}$ (K) & $E_{\mathrm{clo}}$ (\%)
& $E_{\mathrm{field}}$ (K) & $E_{\mathrm{out}}$ (K) & $E_{\mathrm{clo}}$ (\%)
& $E_{\mathrm{field}}$ (K) & $E_{\mathrm{out}}$ (K) & $E_{\mathrm{clo}}$ (\%) \\
\midrule
E-MPINN & IJHMT'25
& 2.95 & \underline{0.793} & \underline{4.59}
& 4.28 & 9.358 & 7.26
& 5.93 & 7.451 & 8.04
& 6.94 & 10.487 & 239.33 \\
RoPINN-CHT & NeurIPS'24
& 5.49 & 2.366 & 25.36
& 4.34 & 9.265 & 12.75
& 5.63 & 7.394 & 12.63
& 10.99 & 9.600 & 10.20 \\
CoPINN-CHT & ICML'25
& 3.16 & 0.869 & 5.86
& 4.53 & 12.108 & 12.28
& 5.84 & 7.435 & 5.75
& 6.97 & 10.334 & 4092.33 \\
MUSA-PINN-CHT & ICML'26
& \underline{1.75} & 1.117 & 11.88
& \underline{4.20} & \underline{2.108} & \underline{1.29}
& \underline{3.65} & \underline{0.937} & \underline{2.87}
& \underline{5.04} & \underline{2.326} & \underline{5.96} \\
\midrule
\textbf{Cova-PINN} & \textbf{Ours}
& \textbf{1.40} & \textbf{0.316} & \textbf{1.01}
& \textbf{2.96} & \textbf{1.890} & \textbf{0.92}
& \textbf{3.12} & \textbf{0.828} & \textbf{2.73}
& \textbf{2.98} & \textbf{1.006} & \textbf{4.09} \\
\textbf{IMP.} & --
& \textbf{20.0\%} & \textbf{60.2\%} & \textbf{77.9\%}
& \textbf{29.5\%} & \textbf{10.3\%} & \textbf{28.8\%}
& \textbf{14.5\%} & \textbf{11.7\%} & \textbf{4.8\%}
& \textbf{40.9\%} & \textbf{56.7\%} & \textbf{31.3\%} \\
\bottomrule
\end{tabular}}
\caption{\textbf{End-to-end accuracy on four TPMS exchangers.}
We report full-field temperature, outlet bulk-temperature, and device-level closure errors. Lower is better; bold and underline mark the best and second-best results. IMP.\ is Cova-PINN's improvement over the strongest baseline. Exact definitions and heat-duty results are given in Appendix~\suppref{app:metrics} and \suppref{app:results}.}
\label{tab:main-results}
\end{table*}

\section{Experiments}
\label{sec:experiments}

\subsection{Experimental Setup}

\begin{figure}[t]
\centering
\includegraphics[width=\columnwidth]{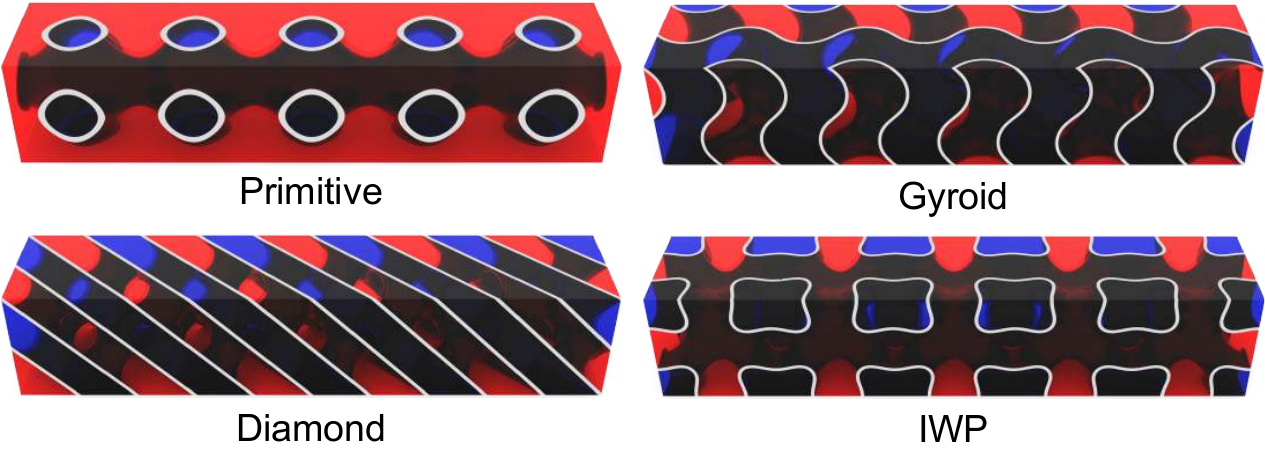}
\caption{\textbf{TPMS exchanger geometries.}
Top: Primitive and Gyroid; bottom: Diamond and IWP. Each exchanger contains five streamwise unit cells. Red and blue denote the hot- and cold-fluid domains, while dark gray denotes the separating solid wall.}
\label{fig:tpms-geometries}
\end{figure}

\paragraph{Problem setting.}
We use Primitive, Gyroid, Diamond, and IWP TPMS exchangers, each containing five streamwise unit cells in $[0,5]\times[0,1]\times[0,1]$, plus a more complex, geometrically distinct DualMS U-shaped exchanger \cite{zhang2025dualms}. Figure~\ref{fig:tpms-geometries} visualizes the coupled three-dimensional domains of the TPMS family. In every TPMS case, both streams enter at $x=0$ on the left and exit at $x=5$ on the right, giving the left-to-right parallel-flow configuration shown in the figure. Their channel connectivity and fluid--solid interface orientation vary substantially across topologies, providing geometrically distinct supports for the same CHT equations and training protocol. Both streams operate at $Re=100$ with inlet temperatures of $80\,^\circ\mathrm C$ and $20\,^\circ\mathrm C$. ANSYS CFX provides reference fields; geometry, boundary conditions, and solver settings are detailed in Appendix~\suppref{app:reference}.

\begin{figure*}[t]
\centering
\includegraphics[width=\textwidth]{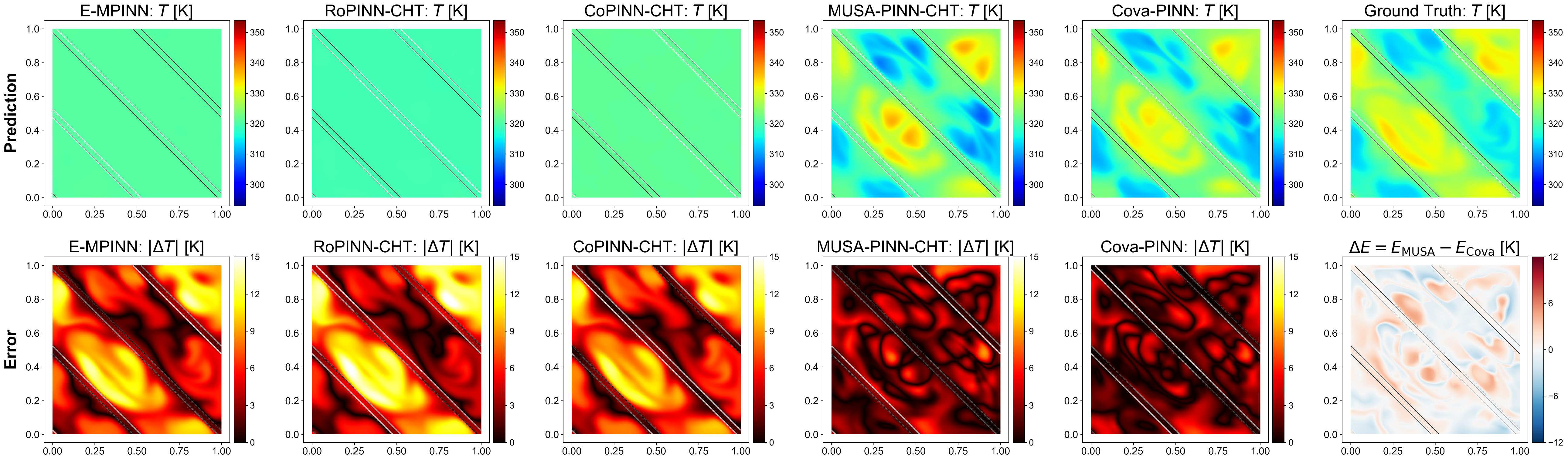}
\caption{\textbf{Outlet-temperature comparison on the Diamond exchanger at $x=5$.}
Top: predicted temperatures from the five methods and CFX; bottom: the corresponding absolute-error fields, followed by
$\Delta E=|T_{\mathrm{MUSA}}-T_{\mathrm{CFX}}|-|T_{\mathrm{Cova}}-T_{\mathrm{CFX}}|$ in the rightmost panel. Prediction and absolute-error panels use matched color ranges. Positive red values in $\Delta E$ indicate lower Cova-PINN error, whereas negative blue values indicate lower MUSA-PINN-CHT error.}
\label{fig:d-outlet}
\end{figure*}

\paragraph{Baselines.}
We compare with the CHT-specific E-MPINN \cite{lee2025extended}, optimization-oriented RoPINN and CoPINN \cite{wu2024ropinn,duan2025copinn}, and geometry-aware MUSA-PINN \cite{zhang2026musa}. MUSA-PINN-CHT uses MUSA-PINN for both flows and the pointwise CHT objective for all three temperatures, making it the closest baseline.

All methods are evaluated with matched network capacity, complete CHT physics, and the same total number of optimization updates. Appendix~\suppref{app:baseline-fairness} details the baseline adaptations, sampling protocol, staged versus joint training, and computational cost.

\paragraph{Implementation details.}
Each subnetwork uses $3{\to}60$ Fourier features \cite{tancik2020fourier} followed by five width-$256$ $\tanh$ layers. Training uses SOAP \cite{wang2025gradient} and PyTorch 2.4.1 with CUDA 12.4 on an NVIDIA A40. Network and optimization settings are listed in Appendix Table~\suppref{tab:sup-training}.

\paragraph{Evaluation metrics.}
We report four complementary metrics. $E_{\mathrm{field}}$ is the pooled temperature MAE over all three media and measures distributed field reconstruction. Because exchanger performance depends on cumulative energy transfer, $E_{\mathrm{out}}$ averages the hot- and cold-stream bulk outlet-temperature errors; common CFX mass-flux weights isolate temperature accuracy from learned-flow differences. A small closure error can nevertheless hide similarly biased heat rates, so $E_Q$ compares the normalized hot-side heat release $Q_h$ and cold-side heat gain $Q_c$ with the CFX reference:
\[
E_Q
=100\sqrt{\frac{
(Q_h^{\mathrm{pred}}-Q_h^{\mathrm{ref}})^2+
(Q_c^{\mathrm{pred}}-Q_c^{\mathrm{ref}})^2}{2}}\%.
\]
Finally, $E_{\mathrm{clo}}=100|Q_h^{\mathrm{pred}}-Q_c^{\mathrm{pred}}|\%$ directly measures device-level energy imbalance. Temperatures are reported in kelvin; $Q_h$ and $Q_c$ are normalized by $Q_0=C_{\min}\Delta T$. Appendix~\suppref{app:metrics} gives the exact estimators.

\subsection{Comparative Analysis on TPMS Exchangers}

\paragraph{Overall performance.}
Table~\ref{tab:main-results} shows that Cova-PINN is best in every reported TPMS cell. Relative to MUSA-PINN-CHT, it reduces average $E_{\mathrm{field}}$, $E_{\mathrm{out}}$, and $E_{\mathrm{clo}}$ by $28.6\%$, $37.7\%$, and $60.2\%$, respectively. The independently evaluated $E_Q$ also falls by $44.8\%$, confirming more accurate exchanged heat rather than closure alone.

\paragraph{Performance across geometries.}
The gains persist across interface topologies: Primitive shows $60.2\%$ lower outlet error and $77.9\%$ lower closure error than the strongest baselines, while on IWP Cova-PINN reduces MUSA-PINN-CHT's field/outlet errors from $5.04/2.326$ to $2.98/1.006$~K. It also achieves the lowest closure on Gyroid and Diamond.

\begin{figure}[t]
\centering
\includegraphics[width=\columnwidth]{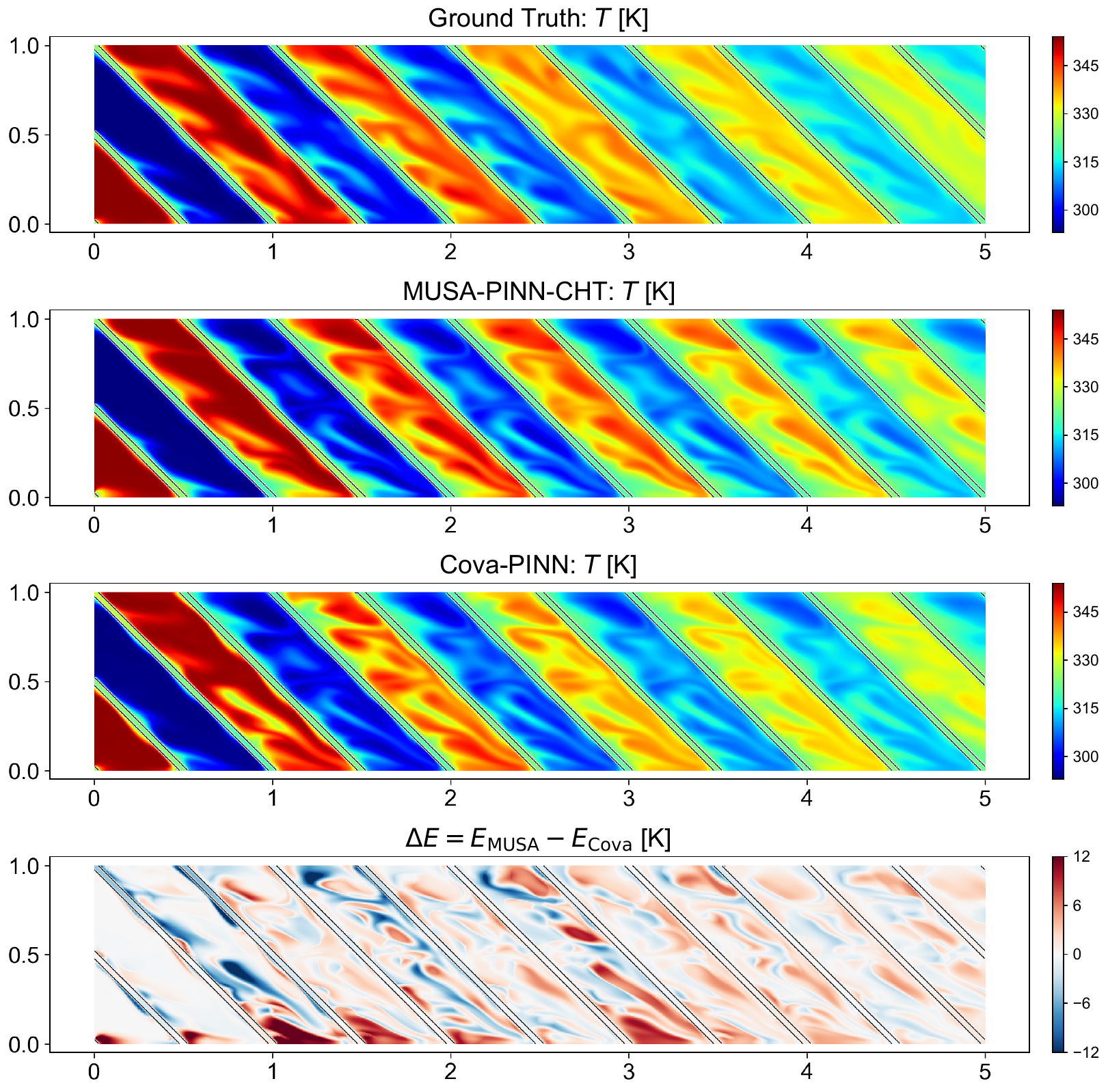}
\caption{\textbf{Longitudinal temperature comparison on the Diamond exchanger at $z=0.5$.}
From top to bottom: CFX, MUSA-PINN-CHT, and Cova-PINN temperatures, followed by
$\Delta E=|T_{\mathrm{MUSA}}-T_{\mathrm{CFX}}|-|T_{\mathrm{Cova}}-T_{\mathrm{CFX}}|$.
The temperature panels share one color range. Positive red values indicate lower Cova-PINN error; negative blue values indicate lower MUSA-PINN-CHT error.}
\label{fig:d-longitudinal}
\end{figure}

Figure~\ref{fig:d-outlet} compares all methods on the Diamond outlet plane $x=5$, complementing the bulk outlet metric with spatially resolved errors. Having established the full-baseline comparison at the outlet, Figure~\ref{fig:d-longitudinal} focuses on MUSA-PINN-CHT, the closest structural baseline, and follows the internal temperature field through all five cells at $z=0.5$. Its signed error-difference map localizes the aggregate improvement: red regions favor Cova-PINN, whereas blue regions favor MUSA-PINN-CHT.
The full longitudinal comparison with all baselines is provided in Appendix ~\suppref{app:d_results}.

\subsection{Cross-Family Evaluation on DualMS}

\begin{figure}[t]
\centering
\includegraphics[width=\columnwidth]{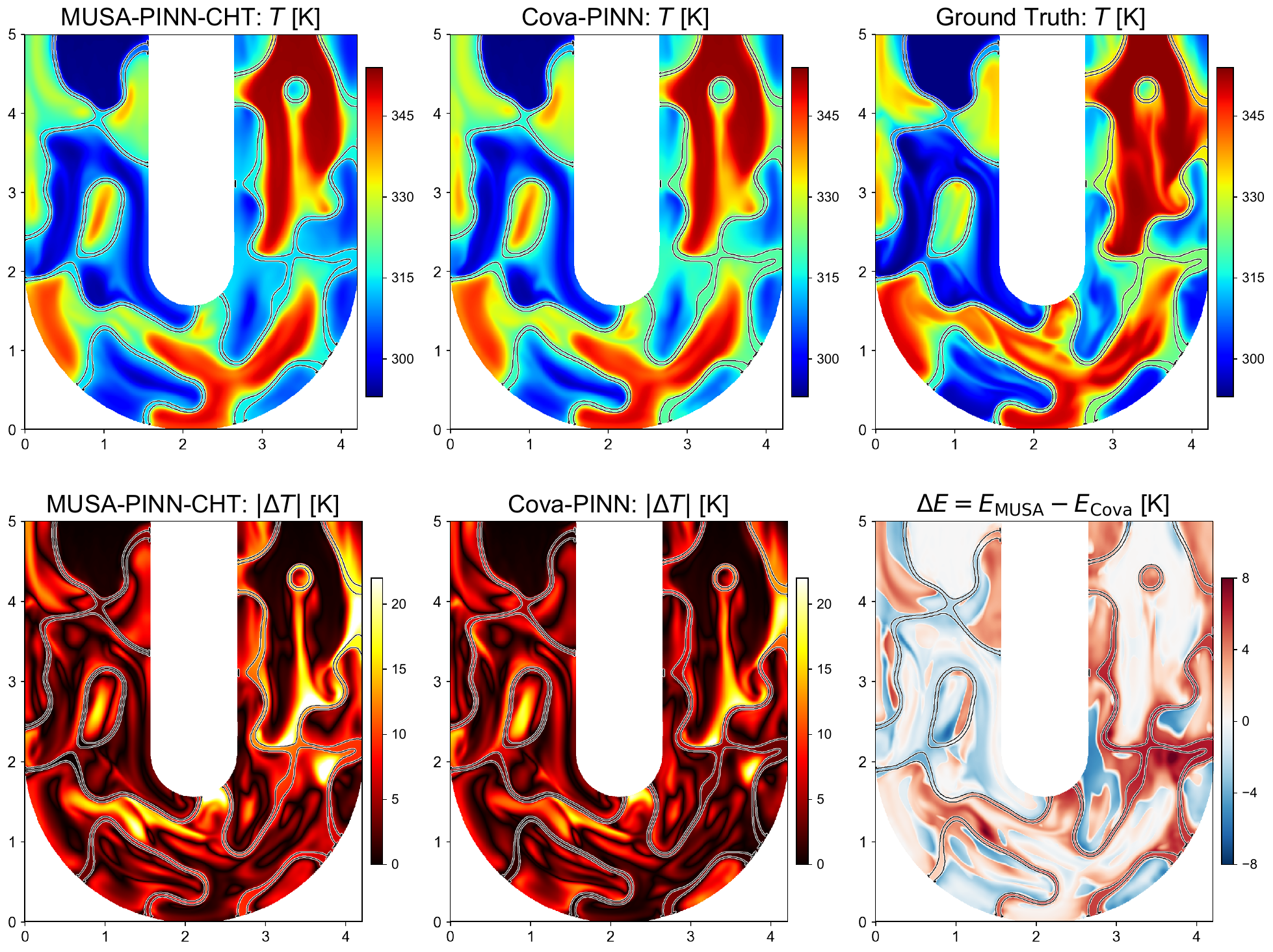}
\caption{\textbf{Temperature field comparison on DualMS.}
All panels show the common mid-plane of the two counter-flow U-shaped passages. Top: MUSA-PINN-CHT, Cova-PINN, and CFX temperatures. Bottom: the two absolute-error fields and
$\Delta E=E_{\mathrm{MUSA}}-E_{\mathrm{Cova}}$; positive red values indicate lower Cova-PINN error, whereas negative blue values indicate lower MUSA-PINN-CHT error.}
\label{fig:dualms-longitudinal}
\end{figure}

To test Cova-PINN on a more complex geometry outside the periodic TPMS family, we use the DualMS exchanger, in which the hot and cold streams traverse separate U-shaped passages in opposite directions. This counter-flow arrangement changes both interface topology and transport pathway; Figure~\ref{fig:dualms-longitudinal} displays the common mid-plane of the two passages (Table~\ref{tab:dualms}).

\begin{table}[t]
\centering

\scriptsize
\setlength{\tabcolsep}{2.8pt}
\resizebox{\columnwidth}{!}{%
\begin{tabular}{lrrrr}
\toprule
\textbf{Method} & $\boldsymbol{E_{\mathrm{field}}}$ \textbf{(K)}
& $\boldsymbol{E_{\mathrm{out}}}$ \textbf{(K)}
& $\boldsymbol{E_Q}$ \textbf{(\%)} & $\boldsymbol{E_{\mathrm{clo}}}$ \textbf{(\%)} \\
\midrule
MUSA-PINN-CHT & 5.996 & 4.250 & 7.037 & 13.575 \\
\textbf{Cova-PINN} & \textbf{4.802} & \textbf{1.354} & \textbf{1.998} & \textbf{3.590} \\
\bottomrule
\end{tabular}}
\caption{\textbf{Cross-family evaluation on the geometrically distinct DualMS exchanger.}
Cova-PINN is compared with MUSA-PINN-CHT, its closest complex-geometry baseline, using field, outlet, heat-duty, and closure errors. Lower is better for all metrics.}
\label{tab:dualms}
\end{table}

Cova-PINN follows CFX more closely along the curved pathway and reduces $E_{\mathrm{field}}$, $E_{\mathrm{out}}$, $E_Q$, and $E_{\mathrm{clo}}$ by $19.9\%$, $68.1\%$, $71.6\%$, and $73.6\%$, respectively. The larger exchanger-level gains show that transfer fidelity warrants evaluation beyond field MAE.

\begin{table}[!h]
\centering

\scriptsize
\setlength{\tabcolsep}{3.5pt}
\resizebox{\columnwidth}{!}{%
\begin{tabular}{lccc}
\toprule
\textbf{Variant} & $E_{\mathrm{field}}$ (K) & $E_{\mathrm{out}}$ (K) & $E_{\mathrm{clo}}$ (\%) \\
\midrule
\textbf{Full Cova-PINN} & \textbf{2.98} & \textbf{1.006} & \textbf{4.09} \\
w/o Local CV (L) & 4.81 & 2.157 & 6.59 \\
w/o Global Closure (G) & 5.00 & 2.335 & 7.32 \\
w/o Solid-Mean Reg. (R) & 4.56 & 4.568 & 23.87 \\
\bottomrule
\end{tabular}}
\caption{\textbf{Component-wise ablation of Cova-PINN.}
The controlled study uses the IWP benchmark throughout. Starting from full Cova-PINN, each variant removes local composite-control-volume conservation (L), global paired-wall heat-rate closure (G), or annealed solid-mean regularization (R), while keeping all other training settings fixed. Lower is better for all metrics.}
\label{tab:ablations}
\end{table}

\subsection{Ablation Studies}
\label{sec:ablation}

We isolate the roles of the finite-region, device-scale, and early-stage objectives through a controlled leave-one-component-out study on a fixed TPMS benchmark.

Table~\ref{tab:ablations} shows that every removal degrades all three reported metrics. Removing G gives the largest field error ($5.00$~K), whereas removing R most strongly harms outlet and closure accuracy ($4.568$~K and $23.87\%$). $E_Q$ likewise rises from $4.21\%$ to $4.36\%$, $4.75\%$, and $12.71\%$ without L, G, and R, respectively. These distinct degradation patterns confirm that the finite-region, device-scale, and early-stage objectives make complementary contributions.

\section{Discussion and Conclusion}

In this paper, we presented Cova-PINN, a support-aligned multi-network framework for fluid--solid CHT in complex geometries. Its global--local conservation co-optimization coordinates finite-region transfer through wall-centered composite control volumes and exchanger-scale transfer through paired-wall closure, while annealed solid-mean regularization conditions early training. Relative to MUSA-PINN-CHT, it reduces average field, outlet, heat-duty, and closure errors across four TPMS topologies by $28.6\%$, $37.7\%$, $44.8\%$, and $60.2\%$, respectively. On the more complex DualMS exchanger, the corresponding reductions are $19.9\%$, $68.1\%$, $71.6\%$, and $73.6\%$. Component ablations further verify the complementary roles of all three objectives. These results establish conservation-support alignment as an effective principle for multi-domain physics learning.

\textbf{Limitations.}
The present study considers steady, constant-property CHT with temperature-independent flow fields; transient transport and temperature-dependent flow--thermal feedback remain outside its scope. Local-control-volume and paired-wall quadrature also increase training cost (Appendix~\suppref{app:cost}).

\textbf{Future Work.}
Future work will extend support-aligned conservation to coupled and transient thermofluid problems, learn adaptive control-volume placement from geometry and residuals, and develop sparse quadrature. Parameterized studies across operating conditions and material contrasts will assess transferability.

\bibliography{aaai2027}

\ifwithappendix
\clearpage
\onecolumn
\appendix
\setcounter{secnumdepth}{2}
\appendixtableofcontents
\clearpage
\twocolumn
\enableappendixtoc
\section{Complete Problem Specification}
\label{app:problem}

\subsection{Domains, Assumptions, and Nondimensionalization}

Let $\Omega_h$, $\Omega_s$, and $\Omega_c\subset\mathbb R^3$ be the mutually disjoint interiors of the hot-fluid, solid, and cold-fluid regions. Their conjugate interfaces are $\Gamma_{hs}=\partial\Omega_h\cap\partial\Omega_s$ and $\Gamma_{sc}=\partial\Omega_s\cap\partial\Omega_c$. We consider steady incompressible forced convection with constant material properties, negligible buoyancy, and no volumetric heat sources. The hot and cold streams use the same fluid with properties $\rho_f$, $c_{p,f}$, $\mu_f$, and $k_f$. Temperature does not feed back into momentum, so the two flow fields are learned first and frozen throughout thermal training.

Dimensional quantities carry the superscript ${\rm d}$, whereas symbols without this superscript are dimensionless. With geometric coordinate scale $L_0$, inlet mean speed $U_{0,f}$ for fluid $f\in\{h,c\}$, and $\Delta T=T_{h,\mathrm{in}}^{\rm d}-T_{c,\mathrm{in}}^{\rm d}$, the variables are nondimensionalized as
\begin{equation}
\begin{aligned}
\mathbf x&=\frac{\mathbf x^{\rm d}}{L_0},&
\mathbf u_f&=\frac{\mathbf u_f^{\rm d}}{U_{0,f}},\\
p_f&=\frac{p_f^{\rm d}-p_{f,\mathrm{out}}^{\rm d}}{\rho_fU_{0,f}^2},&
T_m&=\frac{T_m^{\rm d}-T_{c,\mathrm{in}}^{\rm d}}{\Delta T},
\end{aligned}
\label{eq:sup-nondim}
\end{equation}
where $m\in\{h,s,c\}$. Let $C_f=\dot m_fc_{p,f}$, $C_{\min}=\min(C_h,C_c)$, and $Q_0=C_{\min}\Delta T$. The dimensionless flow groups and conductivity ratios are
\begin{equation}
\begin{aligned}
Re_f&=\frac{\rho_fU_{0,f}L_0}{\mu_f},&
Pe_f&=\frac{\rho_fc_{p,f}U_{0,f}L_0}{k_f},\\
\kappa_m&=\frac{k_m}{k_f},\quad m\in\{h,s,c\},&
\kappa_h&=\kappa_c=1.
\end{aligned}
\label{eq:sup-groups}
\end{equation}

The characteristic length is the inlet hydraulic diameter,
\begin{equation}
L_0=D_h=\frac{4A_{\mathrm{in}}}{P_{\mathrm{in}}},
\label{eq:sup-hydraulic-diameter}
\end{equation}
where $A_{\mathrm{in}}$ and $P_{\mathrm{in}}$ are the area and wetted perimeter of the inlet cross-section. The topology-specific dimensional values of $L_0$ are reported in Table~\ref{tab:sup-benchmark-config}.

\subsection{Governing Equations and Interface Conditions}

The dimensionless flow equations in each fluid are
\begin{equation}
\begin{aligned}
\nabla\!\cdot\mathbf u_f&=0,\\
(\mathbf u_f\!\cdot\nabla)\mathbf u_f+\nabla p_f
-\frac{1}{Re_f}\nabla^2\mathbf u_f&=\mathbf0,
\qquad f\in\{h,c\}.
\end{aligned}
\label{eq:sup-flow}
\end{equation}
With no internal heat generation, the energy equations are
\begin{equation}
\begin{aligned}
\mathbf u_f\!\cdot\nabla T_f-\frac{1}{Pe_f}\nabla^2T_f&=0,
&f&\in\{h,c\},\\
\nabla^2T_s&=0.&&
\end{aligned}
\label{eq:sup-energy}
\end{equation}
For integral balances across media, all fluxes are expressed using the common reference heat-flux scale $k_f\Delta T/L_0$:
\begin{equation}
\begin{aligned}
\mathbf J_f&=Pe_f\mathbf u_fT_f-\nabla T_f,
&&f\in\{h,c\},\\
\mathbf J_s&=-\kappa_s\nabla T_s,\qquad
&&\nabla\!\cdot\mathbf J_m=0,\quad m\in\{h,s,c\}.
\end{aligned}
\label{eq:sup-common-flux}
\end{equation}
Their surface integrals therefore share the heat-rate scale $Q_{\mathrm{ref}}=k_f\Delta T L_0$. The reporting scale $Q_0$ is used separately for the outlet-based energy-imbalance metric in Sec.~\ref{app:metrics}.
Define $\Gamma_h:=\Gamma_{hs}$ and $\Gamma_c:=\Gamma_{sc}$. If $\mathbf n_f$ and $\mathbf n_s$ point outward from the adjoining fluid and solid regions, respectively, the conjugate conditions on $\Gamma_f$, for $f\in\{h,c\}$, are
\begin{equation}
\mathbf u_f=\mathbf0,\qquad
T_f=T_s,\qquad
\nabla T_f\!\cdot\mathbf n_f+\kappa_s\nabla T_s\!\cdot\mathbf n_s=0.
\label{eq:sup-interface}
\end{equation}

\subsection{Boundary and Material Configuration}

Table~\ref{tab:sup-bc} summarizes the boundary conditions. The phrase ``no slip'' applies to fluid velocity on a fluid--solid wall; the solid does not have a velocity field.

\begin{table}[t]
\centering

\scriptsize
\setlength{\tabcolsep}{3.5pt}
\resizebox{\columnwidth}{!}{%
\begin{tabular}{lll}
\toprule
\textbf{Boundary} & \textbf{Flow condition} & \textbf{Thermal condition} \\
\midrule
Hot inlet & Uniform, $|\mathbf u_h^{\rm d}|=1\,\mathrm{m\,s^{-1}}$ along $+x$ & $T_h^{\rm d}=80\,^\circ\mathrm C$ \\
Cold inlet & Uniform, $|\mathbf u_c^{\rm d}|=1\,\mathrm{m\,s^{-1}}$ along $+x$ & $T_c^{\rm d}=20\,^\circ\mathrm C$ \\
Fluid outlets & Mean gauge pressure $\overline p_f^{\rm d}=0\,\mathrm{Pa}$ & No direct thermal constraint \\
$\Gamma_{hs},\Gamma_{sc}$ & No slip & Temperature and normal-flux continuity \\
External solid boundary & -- & Adiabatic \\
\bottomrule
\end{tabular}}
\caption{\textbf{Boundary conditions for the four TPMS benchmarks.}
All methods use the same flow and thermal conditions; no direct thermal constraint is imposed at the fluid outlets.}
\label{tab:sup-bc}
\end{table}

For DualMS, the same inlet speed magnitudes and temperatures, outlet gauge pressure, interface conditions, and external adiabatic condition are applied to its labeled ports and walls; inlet velocity directions follow the inward normals of the corresponding ports.

The four principal TPMS cases use $Re_h=Re_c=100$, $T_{h,\mathrm{in}}^{\rm d}=80\,^\circ\mathrm C$, and $T_{c,\mathrm{in}}^{\rm d}=20\,^\circ\mathrm C$, hence $\Delta T=60\,\mathrm K$. The hot and cold streams have identical constant properties within each benchmark. Density, heat capacity, and conductivity are shared across topologies and are listed in Table~\ref{tab:sup-physics}. Because the inlet hydraulic diameter $L_0$ is topology dependent, the viscosity is set as $\mu_f=\rho_fU_{0,f}L_0/Re_f$ to retain $Re_f=100$; the resulting $L_0$, $\mu_f$, and $Pe_f$ values are reported in Table~\ref{tab:sup-benchmark-config}.

\begin{table}[t]
\centering

\scriptsize
\setlength{\tabcolsep}{4pt}
\begin{tabular}{ll}
\toprule
\textbf{Quantity} & \textbf{Value} \\
\midrule
$Re_h/Re_c$ & $100/100$ \\
$T_{h,\mathrm{in}}^{\rm d}/T_{c,\mathrm{in}}^{\rm d}$ & $80/20\,^\circ\mathrm C$ \\
$U_{0,h}^{\rm d}/U_{0,c}^{\rm d}$ & $1/1\,\mathrm{m\,s^{-1}}$ \\
$\rho_f$ & $1\,\mathrm{kg\,m^{-3}}$ \\
$c_{p,f}$ & $1000\,\mathrm{J\,kg^{-1}\,K^{-1}}$ \\
$k_f$ & $0.6069\,\mathrm{W\,m^{-1}\,K^{-1}}$ \\
$k_s$ & $237\,\mathrm{W\,m^{-1}\,K^{-1}}$ \\
$\kappa_s=k_s/k_f$ & $390.509$ \\
$L_0,\mu_f,Pe_f$ & Topology-specific; Table~\ref{tab:sup-benchmark-config} \\
$Q_{\mathrm{ref}}$ & $k_f\Delta T L_0$ \\
$Q_0$ & $C_{\min}\Delta T$ \\
\bottomrule
\end{tabular}
\caption{\textbf{Shared operating conditions and thermal material parameters for the TPMS benchmarks.}
Both streams use the listed values in every topology; topology-specific $L_0$, $\mu_f$, and $Pe_f$ are reported in Table~\ref{tab:sup-benchmark-config}.}
\label{tab:sup-physics}
\end{table}

\section{Geometry Construction}
\label{app:geometry}

\subsection{TPMS Benchmark Family}

Primitive (P), Gyroid (G), Diamond (D), and IWP form the main benchmark family. Let $\mathbf x=(x,y,z)$ denote the dimensionless coordinates used to construct the unit cells. Each geometry contains five unit cells tiled along $x$ and is constructed in
\begin{equation}
\mathcal D=[0,5]\times[0,1]\times[0,1].
\label{eq:sup-box}
\end{equation}
Both streams enter at $x=0$ and move along the positive $x$ direction toward outlets at $x=5$, giving a parallel-flow configuration. Table~\ref{tab:sup-benchmark-config} reports the topology-specific hydraulic diameters and the resulting flow parameters.

\begin{table}[t]
\centering

\scriptsize
\setlength{\tabcolsep}{2.2pt}
\resizebox{\columnwidth}{!}{%
\begin{tabular}{lccccc}
\toprule
\textbf{Geometry} & \textbf{Array} & $\boldsymbol{L_0}$ \textbf{(m)}
& $\boldsymbol{Re_h=Re_c}$ & $\boldsymbol{\mu_f}$ \textbf{(Pa\,s)}
& $\boldsymbol{Pe_f}$ \\
\midrule
Primitive (P) & $5\times1\times1$ & $0.5220$ & $100$ & $5.220{\times}10^{-3}$ & $860.109$ \\
Gyroid (G) & $5\times1\times1$ & $0.4490$ & $100$ & $4.490{\times}10^{-3}$ & $739.825$ \\
Diamond (D) & $5\times1\times1$ & $0.3910$ & $100$ & $3.910{\times}10^{-3}$ & $644.258$ \\
IWP & $5\times1\times1$ & $0.4356$ & $100$ & $4.356{\times}10^{-3}$ & $717.746$ \\
\bottomrule
\end{tabular}}
\caption{\textbf{Topology-specific reference lengths and dimensionless flow parameters for the TPMS benchmarks.}
Here $L_0=4A_{\mathrm{in}}/P_{\mathrm{in}}$. With $\rho_f=1\,\mathrm{kg\,m^{-3}}$ and $U_{0,f}=1\,\mathrm{m\,s^{-1}}$, the listed viscosity gives $Re_h=Re_c=100$, and $Pe_f=\rho_fc_{p,f}U_{0,f}L_0/k_f$.}
\label{tab:sup-benchmark-config}
\end{table}

The geometry source uses the following periodic implicit functions. For compactness, let $c_x=\cos(2\pi x)$ and $s_x=\sin(2\pi x)$, with cyclic definitions for $y,z$, and let $c_{2x}=\cos(4\pi x)$.
\begin{align}
\Phi_P&=c_x+c_y+c_z,\label{eq:sup-p}\\
\Phi_G&=s_xc_y+s_yc_z+s_zc_x,\label{eq:sup-g}\\
\Phi_D&=s_xs_ys_z+s_xc_yc_z+c_xs_yc_z+c_xc_ys_z,\label{eq:sup-d}\\
\Phi_{\mathrm{IWP}}&=2(c_xc_y+c_yc_z+c_zc_x)
-(c_{2x}+c_{2y}+c_{2z}).\label{eq:sup-iwp}
\end{align}
Equations~\eqref{eq:sup-p}--\eqref{eq:sup-iwp} reproduce the four implicit functions used to construct the benchmark geometries. The same nominal wall thickness, $0.03$ in normalized coordinates, is used for all four topologies. The two connected fluid regions produced by the sheet construction are assigned to the hot and cold streams before volume meshing. These geometries are treated as distinct benchmark domains; the experiments do not compare their intrinsic heat-exchanger performance against one another.

\subsection{DualMS Geometry}

DualMS \cite{zhang2025dualms} is used as a geometrically distinct cross-family case. We adopt the fixed U-shaped design generated with the 300-point design configuration and use the resulting watertight solid and two fluid volumes directly for sampling and CFX meshing. Every method is trained independently on this geometry; no geometry optimization is performed as part of the present study.

\section{Spherical Composite Control Volumes}
\label{app:local-cv}

\subsection{Composite Outer-Boundary Construction}

Let $\Gamma_{\mathrm w}:=\Gamma_{hs}\cup\Gamma_{sc}$ denote the two conjugate heat-transfer walls. We sample the center $\mathbf c_j$ on $\Gamma_{\mathrm w}$ so that candidate parent regions are preferentially placed along the thermal interaction path. A fixed-radius spherical parent region is
\begin{equation}
\begin{aligned}
B_j&=B(\mathbf c_j,r)
=\{\mathbf x:\|\mathbf x-\mathbf c_j\|_2\leq r\},\\
\mathbf c_j&\in\Gamma_{\mathrm w},
\qquad r=0.05.
\end{aligned}
\label{eq:sup-ball}
\end{equation}
Let $\Omega$ denote the complete device domain with $\overline{\Omega}=\overline{\Omega}_h\cup\overline{\Omega}_s\cup\overline{\Omega}_c$. Intersecting the parent region with each medium gives
\begin{equation}
V_{m,j}=B_j\cap\Omega_m,\qquad
V_j=\bigcup_{m\in\{h,s,c\}}V_{m,j}=B_j\cap\Omega.
\label{eq:sup-material-portions}
\end{equation}
At each refresh, 50 centers are sampled randomly from the available solid-wall samples. The radius $0.05$ exceeds the nominal wall thickness $0.03$, so the resulting composite supports extend across adjoining material regions along the thermal interaction path. The exact configuration is summarized in Table~\ref{tab:sup-cv-config}.

The sampled boundary is the outer boundary of the composite volume, partitioned into material-labeled patches
\begin{equation}
\begin{aligned}
S_{m,j}&=\partial V_j\cap\overline{\Omega}_m,
&&m\in\{h,s,c\},\\
\partial V_j&=S_{h,j}\cup S_{s,j}\cup S_{c,j}.
\end{aligned}
\label{eq:sup-outer-partition}
\end{equation}
The partition is understood up to measure-zero intersection curves. The internal patches $B_j\cap\Gamma_{hs}$ and $B_j\cap\Gamma_{sc}$ are not included in $S_{m,j}$ and are never sampled by the local composite-control-volume loss. If a ball lies inside the device envelope, $\partial V_j=\partial B_j$ and each nonempty $S_{m,j}$ is a material-labeled spherical patch. A ball that reaches an inlet, outlet, wall, or other exterior device boundary is not rejected: the reached physical-boundary patches are retained and integrated in the same way as the other components of $\partial V_j$.

On a spherical outer patch, the outward unit normal is analytic:
\begin{equation}
\mathbf n_j^{\mathrm{sph}}(\mathbf x)=\frac{\mathbf x-\mathbf c_j}{r}.
\label{eq:sup-sphere-normal}
\end{equation}
On a reached physical-boundary patch, the corresponding outward geometry normal is used.

\subsection{Area-Uniform Sampling and Total-Area Weighting}

For each parent ball, $N_B=10{,}000$ candidate boundary points are sampled and classified by their material and boundary labels. No candidates are drawn from the internal interfaces $\Gamma_{hs}$ or $\Gamma_{sc}$. Each nonempty outer-boundary component $\mathcal S\subset\partial V_j$ with area-uniform samples $\{\mathbf x_k\}_{k=1}^{K_{\mathcal S}}$ is evaluated as
\begin{equation}
\widehat I(\mathcal S)=\frac{\widehat A_{\mathcal S}}{K_{\mathcal S}}
\sum_{k=1}^{K_{\mathcal S}}g_m(\mathbf x_k).
\label{eq:sup-mc-component}
\end{equation}
Here $\widehat A_{\mathcal S}$ is the total area of the component. Every accepted point within one component has equal weight. Only the component's total area multiplies the sample mean; no triangle-wise or pointwise local-area factor is used. Estimates from separately sampled components are added only after this total-area weighting.

For the local energy constraint, $g_m=\mathbf J_m\!\cdot\mathbf n_j$ and
\begin{equation}
\begin{aligned}
\widehat R_j^{\mathrm{cv}}
&=\sum_{\mathcal S\in\mathcal P_j}\widehat I(\mathcal S),\\
\mathcal L_{\mathrm{local}}
&=\frac{1}{N_{\mathrm{cv}}}
\sum_{j=1}^{N_{\mathrm{cv}}}\left(\widehat R_j^{\mathrm{cv}}\right)^2,
\qquad N_{\mathrm{cv}}=50.
\end{aligned}
\label{eq:sup-local-loss}
\end{equation}
Here $\mathcal P_j$ is the set of nonempty, material-labeled outer-boundary components of $V_j$, including any reached physical exterior patch but excluding internal fluid--solid interfaces. The implementation uses one radius and therefore requires no scale aggregation or scale weights. Centers and candidate samples are regenerated every 500 thermal updates.

\begin{table}[t]
\centering

\scriptsize
\setlength{\tabcolsep}{4pt}
\begin{tabular}{ll}
\toprule
\textbf{Item} & \textbf{Value} \\
\midrule
Center support & $\Gamma_{\mathrm w}=\Gamma_{hs}\cup\Gamma_{sc}$ \\
Center distribution & Random sampling from solid-wall samples \\
Fixed radius $r$ & $0.05$ \\
Training control volumes $N_{\mathrm{cv}}$ & $50$ \\
Candidate points $N_B$ & $10{,}000$ per ball \\
Exterior-boundary handling & Retain and integrate reached patches \\
Internal-interface candidates & None for $\mathcal L_{\mathrm{local}}$ \\
Sample refresh frequency & Every 500 thermal updates \\
Local-loss weight & $10$ \\
Scale weights & None (single radius) \\
\bottomrule
\end{tabular}
\caption{\textbf{Sampling and quadrature configuration for local composite control volumes.}
The same settings are used in all Cova-PINN experiments; the single fixed radius requires no scale aggregation.}
\label{tab:sup-cv-config}
\end{table}

\subsection{Interface Cancellation}

For the derivation only, define $V_{m,j}=V_j\cap\Omega_m$. Integrating the differential conservation law over each material portion, applying the divergence theorem, and summing gives
\begin{equation}
\begin{aligned}
0
={}&\sum_{m\in\{h,s,c\}}
\int_{V_{m,j}}\nabla\!\cdot\mathbf J_m\,\mathrm dV\\
={}&\sum_{m\in\{h,s,c\}}
\int_{\partial V_{m,j}}\mathbf J_m\!\cdot\mathbf n_{m,j}\,\mathrm dA\\
={}&\sum_{m\in\{h,s,c\}}
\int_{S_{m,j}}\mathbf J_m\!\cdot\mathbf n_j\,\mathrm dA\\
&+\int_{\Gamma_{hs}\cap V_j}
\left(\mathbf J_h\!\cdot\mathbf n_h
+\mathbf J_s\!\cdot\mathbf n_s^h\right)\mathrm dA\\
&+\int_{\Gamma_{sc}\cap V_j}
\left(\mathbf J_s\!\cdot\mathbf n_s^c
+\mathbf J_c\!\cdot\mathbf n_c\right)\mathrm dA .
\end{aligned}
\label{eq:sup-interface-cancel}
\end{equation}
No penetration eliminates advective transport through each fluid--solid interface, and the normal conductive-flux condition makes the final two integrals vanish for an exact CHT solution. The remaining first term is precisely the outer-boundary residual used for training. Thus the local loss enforces a finite-region consequence of the coupled PDE and interface system; it does not directly evaluate an interface-flux jump. The separate pointwise temperature- and flux-continuity penalties remain essential because an outer balance alone cannot exclude compensating interface errors.

\section{Paired-Wall Heat-Rate Estimator}
\label{app:global-closure}

The global closure term evaluates both interface heat rates from the solid network using $\mathbf J_s(\mathbf x;\phi_s)=-\kappa_s\nabla T_s(\mathbf x;\phi_s)$. Let $\mathbf n_s^h$ and $\mathbf n_s^c$ point outward from the solid toward the hot and cold fluids. With area-uniform samples $\{\mathbf x_{hs,k}\}_{k=1}^{N_{hs}}\subset\Gamma_{hs}$ and $\{\mathbf x_{sc,k}\}_{k=1}^{N_{sc}}\subset\Gamma_{sc}$, the two full-interface integrals are estimated by
\begin{align}
\widehat Q_{h\to s}(\phi_s)
&=-\frac{|\Gamma_{hs}|}{N_{hs}}
\sum_{k=1}^{N_{hs}}
\mathbf J_s(\mathbf x_{hs,k};\phi_s)\!\cdot\!
\mathbf n_s^h(\mathbf x_{hs,k}),\label{eq:sup-q-hs}\\
\widehat Q_{s\to c}(\phi_s)
&=\frac{|\Gamma_{sc}|}{N_{sc}}
\sum_{k=1}^{N_{sc}}
\mathbf J_s(\mathbf x_{sc,k};\phi_s)\!\cdot\!
\mathbf n_s^c(\mathbf x_{sc,k}).\label{eq:sup-q-sc}
\end{align}
Each sample has equal weight within its interface, and the sample mean is multiplied by the corresponding total interface area; no triangle-wise or pointwise local-area factor is used. The resulting residual and loss are
\begin{equation}
\begin{aligned}
R^{\mathrm{wall}}(\phi_s)
&=\widehat Q_{h\to s}(\phi_s)
-\widehat Q_{s\to c}(\phi_s),\\
\mathcal L_{\mathrm{global}}
&=\left(R^{\mathrm{wall}}(\phi_s)\right)^2.
\end{aligned}
\label{eq:sup-global-loss}
\end{equation}
Both estimators depend directly only on $\phi_s$. The fluid temperature networks receive their influence indirectly through the pointwise temperature- and normal-flux-continuity losses on $\Gamma_{hs}$ and $\Gamma_{sc}$. The implementation uses a total of $N_{hs}+N_{sc}=120{,}000$ area-uniform samples across the two interfaces.

\section{Reference Solutions}
\label{app:reference}

\subsection{Reference Solver}

Reference temperature and flow fields are generated using ANSYS CFX 2023 R1. We use its coupled finite-volume solver on unstructured tetrahedral meshes and the High Resolution advection scheme. Five prismatic inflation layers are placed next to the walls, with first-layer height $0.002$ and growth rate $1.2$. The global element size is $0.02$ and is refined to a maximum of $0.01$ near the boundaries, giving approximately $10^7$ volume elements for each TPMS case. Straight inlet and outlet extensions of three TPMS unit-cell lengths are included. Iterations continue until the RMS equation residuals fall below $10^{-8}$. Material properties and boundary conditions follow Tables~\ref{tab:sup-physics} and \ref{tab:sup-bc}. Reference fields are used only for post-training evaluation and do not affect network optimization or any training decision.

\appendixtoccolumnbreak
\section{Experimental Configuration}
\label{app:configuration}

\subsection{Networks, Base Losses, and Sequential Training}

Two independent networks $\mathcal N_h^{\mathrm{flow}}(\mathbf x;\psi_h)$ and $\mathcal N_c^{\mathrm{flow}}(\mathbf x;\psi_c)$ represent $(\mathbf u_h,p_h)$ and $(\mathbf u_c,p_c)$, while $\mathcal N_m^{\mathrm{temp}}(\mathbf x;\phi_m)$ represents the temperature in each medium $m\in\{h,s,c\}$. Let $\boldsymbol{\phi}:=(\phi_h,\phi_s,\phi_c)$ collect the three temperature-network parameter vectors. Each method uses the same field decomposition and complete CHT boundary/interface conditions. Cova first trains the two flow networks using MUSA-PINN and then freezes them; only $\boldsymbol{\phi}$ is updated in the thermal stage. The flow and thermal objectives below are therefore optimized sequentially, not added into one simultaneous loss.

\paragraph{Flow-stage base terms.}
For each $f\in\{h,c\}$, define the continuity and momentum residuals
\begin{equation}
r_f^{\mathrm{div}}=\nabla\!\cdot\mathbf u_f,
\qquad
\mathbf r_f^{\mathrm{mom}}=(\mathbf u_f\!\cdot\nabla)\mathbf u_f+\nabla p_f-\frac{1}{Re_f}\nabla^2\mathbf u_f.
\label{eq:sup-flow-residuals}
\end{equation}
Let $\mathcal X_f^u\subset\Omega_f$ contain $N_f^u$ interior points. The standard strong-form flow loss is
\begin{equation}
\begin{aligned}
\mathcal L_{\mathrm{PDE}}^u
={}&\sum_{f\in\{h,c\}}\frac{\lambda_{\mathrm{div}}}{N_f^u}
\sum_{\mathbf x\in\mathcal X_f^u}|r_f^{\mathrm{div}}(\mathbf x)|^2\\
&+\sum_{f\in\{h,c\}}\frac{\lambda_{\mathrm{mom}}}{N_f^u}
\sum_{\mathbf x\in\mathcal X_f^u}\|\mathbf r_f^{\mathrm{mom}}(\mathbf x)\|_2^2.
\end{aligned}
\label{eq:sup-flow-pde-loss}
\end{equation}
For the prescribed-inlet-velocity, mean-outlet-pressure, and no-slip-wall realization used by the MUSA-based flow stage, let $\mathcal X_{f,\mathrm{in}}^u$, $\mathcal X_{f,\mathrm{out}}^u$, and $\mathcal X_{f,\mathrm w}^u$ be the corresponding boundary samples; $\mathcal X_{f,\mathrm w}^u$ includes the fluid--solid interface and every other no-slip fluid wall. Because the pressure scaling in Eq.~\eqref{eq:sup-nondim} sets the prescribed dimensionless mean outlet pressure to zero, the boundary loss is
\begin{equation}
\begin{aligned}
\mathcal L_{\mathrm{bc}}^u
={}&\sum_{f\in\{h,c\}}
\left(\lambda_{\mathrm{in}}^u\mathcal L_{f,\mathrm{in}}^u
+\lambda_{\mathrm{out}}^u\mathcal L_{f,\mathrm{out}}^u
+\lambda_{\mathrm w}^u\mathcal L_{f,\mathrm w}^u\right),\\
\mathcal L_{f,\mathrm{in}}^u
={}&\frac{1}{N_{f,\mathrm{in}}^u}
\sum_{\mathbf x\in\mathcal X_{f,\mathrm{in}}^u}
\|\mathbf u_f(\mathbf x)-\mathbf u_{f,\mathrm{in}}(\mathbf x)\|_2^2,\\
\mathcal L_{f,\mathrm{out}}^u
={}&\left|
\frac{1}{N_{f,\mathrm{out}}^u}
\sum_{\mathbf x\in\mathcal X_{f,\mathrm{out}}^u}p_f(\mathbf x)
\right|^2,\\
\mathcal L_{f,\mathrm w}^u
={}&\frac{1}{N_{f,\mathrm w}^u}
\sum_{\mathbf x\in\mathcal X_{f,\mathrm w}^u}\|\mathbf u_f(\mathbf x)\|_2^2.
\end{aligned}
\label{eq:sup-flow-bc-loss}
\end{equation}
Here $N_{f,b}^u=|\mathcal X_{f,b}^u|$. MUSA-PINN-CHT and Cova use the flow objective
\begin{equation}
\mathcal L_{\mathrm{flow}}^{\mathrm{MUSA}}(t)
=\mathcal L_{\mathrm{PDE}}^u+\mathcal L_{\mathrm{bc}}^u
+\mathcal L_{\mathrm{wk,div}}
+\gamma(t)\mathcal L_{\mathrm{wk,mom}},
\label{eq:sup-musa-flow-loss}
\end{equation}
where the last two terms are the multiscale weak continuity and momentum losses and $\gamma(t)$ is the stage switch defined by MUSA-PINN \cite{zhang2026musa}. The other end-to-end baselines train their own flow networks with their published mechanisms, as specified below; they do not receive flow weights from another method.

\paragraph{Thermal-stage pointwise terms.}
With the frozen velocities $\bar{\mathbf u}_f$, define
\begin{equation}
\begin{aligned}
r_f^T&=\bar{\mathbf u}_f\!\cdot\nabla T_f-\frac{1}{Pe_f}\nabla^2T_f,
&&f\in\{h,c\},\\
r_s^T&=\nabla^2T_s.
\end{aligned}
\label{eq:sup-thermal-residuals}
\end{equation}
At each thermal update, let $\mathcal X_f^T\subset\Omega_f$ and $\mathcal X_s^T\subset\Omega_s$ be the finite energy-equation collocation sets. With $N_f^T=|\mathcal X_f^T|$ and $N_s^T=|\mathcal X_s^T|$, the strong-form energy loss is
\begin{equation}
\begin{aligned}
\mathcal L_{\mathrm{PDE}}^T
={}&\sum_{f\in\{h,c\}}\frac{1}{N_f^T}
\sum_{\mathbf x\in\mathcal X_f^T}|r_f^T(\mathbf x)|^2\\
&+\frac{1}{N_s^T}
\sum_{\mathbf x\in\mathcal X_s^T}|r_s^T(\mathbf x)|^2.
\end{aligned}
\label{eq:sup-pointwise-pde}
\end{equation}
Let $\mathcal B_T$ be the set of non-interface thermal boundary components. For each $\beta\in\mathcal B_T$, let $m(\beta)\in\{h,s,c\}$ denote its medium, $\mathcal X_\beta^T$ its $N_\beta^T$ samples, and $\mathcal B_\beta[T_{m(\beta)}]=g_\beta$ its prescribed boundary operator. Its empirical residual is evaluated explicitly as
\begin{equation}
\mathcal L_{\mathrm{bc}}^T
=\sum_{\beta\in\mathcal B_T}
\frac{\omega_\beta}{N_\beta^T}
\sum_{\mathbf x\in\mathcal X_\beta^T}
\left|
\mathcal B_\beta[T_{m(\beta)}](\mathbf x)-g_\beta(\mathbf x)
\right|^2,
\label{eq:sup-thermal-bc-loss}
\end{equation}
where $\omega_\beta$ is the relative weight of boundary component $\beta$. Under the temperature scaling in Eq.~\eqref{eq:sup-nondim}, the two inlet residuals are $T_h-1$ and $T_c$, respectively. The external-solid residual is $\kappa_s\nabla T_s\!\cdot\mathbf n_s$. Table~\ref{tab:sup-bc} specifies the boundary operator imposed on every component.

For each interface $\Gamma_f$ defined in Sec.~\ref{app:problem}, let $\mathcal X_{\Gamma_f}$ contain $N_{\Gamma_f}$ paired interface points. The empirical interface losses are
\begin{align}
\mathcal L_{\Gamma,T}
&=\sum_{f\in\{h,c\}}\frac{1}{N_{\Gamma_f}}
\sum_{\mathbf x\in\mathcal X_{\Gamma_f}}
|T_f-T_s|^2,\label{eq:sup-interface-t-loss}\\
\mathcal L_{\Gamma,q}
&=\sum_{f\in\{h,c\}}\frac{1}{N_{\Gamma_f}}
\sum_{\mathbf x\in\mathcal X_{\Gamma_f}}
\left|\nabla T_f\!\cdot\mathbf n_f
+\kappa_s\nabla T_s\!\cdot\mathbf n_s\right|^2.
\label{eq:sup-interface-q-loss}
\end{align}
All fields and gradients in Eqs.~\eqref{eq:sup-interface-t-loss}--\eqref{eq:sup-interface-q-loss} are evaluated at the summation point $\mathbf x$. Thus the complete pointwise thermal objective is
\begin{equation}
\mathcal L_{\mathrm{pt}}=
\lambda_{\mathrm{PDE}}\mathcal L_{\mathrm{PDE}}^T
+\lambda_{\mathrm{bc}}\mathcal L_{\mathrm{bc}}^T
+\lambda_{\Gamma,T}\mathcal L_{\Gamma,T}
+\lambda_{\Gamma,q}\mathcal L_{\Gamma,q}.
\label{eq:sup-pointwise-total}
\end{equation}
Each component is normalized by its own sample count before applying the loss weights. The reported network, optimization, and method-specific quadrature settings are summarized in Table~\ref{tab:sup-training}.

\begin{table*}[t]
\centering

\scriptsize
\setlength{\tabcolsep}{4pt}
\begin{tabular}{lll}
\toprule
\textbf{Category} & \textbf{Flow stage} & \textbf{Thermal stage} \\
\midrule
Input/output & $(x,y,z)\mapsto(u,v,w,p)$ & $(x,y,z)\mapsto T_m$ \\
Networks & Two, one per fluid & Three, one per medium \\
Architecture & $3{\to}60$ Fourier features, then $5{\times}256$ $\tanh$ layers & $3{\to}60$ Fourier features, then $5{\times}256$ $\tanh$ layers \\
Input encoding & Linear Fourier frequencies in $[1,2.5]$ \cite{tancik2020fourier} & Linear Fourier frequencies in $[1,2.5]$ \cite{tancik2020fourier} \\
Initialization & Xavier-normal weights; zero biases & Xavier-normal weights; zero biases \\
Optimizer & SOAP \cite{wang2025gradient}, default parameters & SOAP \cite{wang2025gradient}, default parameters \\
Learning rate & $10^{-3}$ & $10^{-3}$ \\
Updates & $5{,}000$ & $5{,}000$ \\
Pointwise loss weights & See Table~\ref{tab:sup-loss-weights} & See Table~\ref{tab:sup-loss-weights} \\
Local CV quadrature & -- & $r=0.05$; 50 balls; $10{,}000$ candidates/ball; refresh every 500 updates \\
Paired-wall quadrature & -- & $120{,}000$ samples across both solid-side interfaces \\
Solid-mean schedule & -- & $N_R=2{,}500$; $\lambda_R(t)=\lambda_R^{(0)}\max(1-t/N_R,0)$ \\
\bottomrule
\end{tabular}
\caption{\textbf{Cova-PINN network, optimization, and conservation-quadrature configuration.}
The flow networks are trained first and frozen before thermal optimization; all subnetworks use the same Fourier-feature MLP backbone.}
\label{tab:sup-training}
\end{table*}

\begin{table*}[t]
\centering

\scriptsize
\setlength{\tabcolsep}{4pt}
\begin{tabular}{lll}
\toprule
\textbf{Stage/objective} & \textbf{Weight(s)} & \textbf{Value(s)} \\
\midrule
Flow strong continuity and momentum
& $\lambda_{\mathrm{div}},\lambda_{\mathrm{mom}}$ & $1,\ 0.1$ \\
Flow inlet, outlet, and no-slip wall
& $\lambda_{\mathrm{in}}^u,\lambda_{\mathrm{out}}^u,\lambda_{\mathrm w}^u$ & $10,\ 10,\ 10$ \\
MUSA weak continuity (large/medium/small)
& $\lambda_{\mathrm{wk,div}}^{L/M/S}$ & $100,\ 100,\ 100$ \\
MUSA weak momentum (large/medium/small)
& $\lambda_{\mathrm{wk,mom}}^{L/M/S}$ & $4,\ 25,\ 100$ \\
Thermal PDE and boundary conditions
& $\lambda_{\mathrm{PDE}},\lambda_{\mathrm{bc}}$ & $1,\ 1$ \\
Interface temperature and heat flux
& $\lambda_{\Gamma,T},\lambda_{\Gamma,q}$ & $1,\ 1$ \\
Local composite control volumes
& $\lambda_L$ & $10$ (fixed) \\
Global paired-wall closure
& $\lambda_G$ & $1$ (fixed) \\
Solid-mean regularization
& $\lambda_R(t)$ & $\lambda_R^{(0)}=1$; linear decay to $0$ by update $2{,}500$ \\
\bottomrule
\end{tabular}
\caption{\textbf{Loss weights used by Cova-PINN and the baseline objectives.}
Strong-form weights are shared across methods; MUSA weak-form flow weights are shared by MUSA-PINN-CHT and Cova-PINN, while the local, global, and regularization weights apply to Cova-PINN.}
\label{tab:sup-loss-weights}
\end{table*}

\subsection{Baseline Selection and Fairness}
\label{app:baseline-fairness}

The end-to-end comparison includes E-MPINN \cite{lee2025extended}, CHT adaptations of RoPINN \cite{wu2024ropinn} and CoPINN \cite{duan2025copinn}, MUSA-PINN-CHT \cite{zhang2026musa}, and \method. E-MPINN provides a CHT-specific multi-network baseline; RoPINN-CHT and CoPINN-CHT represent region-based and curriculum-style PINN optimization; and MUSA-PINN-CHT provides the closest structural baseline for complex-geometry flow. It uses MUSA-PINN for both flow networks and the standard pointwise objective for the three temperature networks.

All methods use the same five-field decomposition, network capacity, complete CHT physics, SOAP with a learning rate of $10^{-3}$, and $10{,}000$ total updates. The baselines jointly optimize their flow and temperature networks for the full schedule. Cova-PINN uses $5{,}000$ flow-stage updates followed by $5{,}000$ thermal-stage updates with the learned flow fields frozen during thermal optimization. The base strong-form collocation sets are identical across methods; MUSA-PINN-CHT and Cova-PINN additionally use the same weak-form fluid sampling configuration. Method-specific operations required by RoPINN, CoPINN, and Cova-PINN are retained. The final iterate is evaluated without early stopping or model selection. Thus the comparison matches total optimization updates rather than wall-clock time; the resulting costs are reported in Table~\ref{tab:sup-cost}.

\section{Evaluation Metrics}
\label{app:metrics}

\subsection{Temperature-Field and Outlet Errors}

Let $\mathcal X_m$ contain the valid uniformly sampled volume-evaluation
points in medium $m\in\{h,s,c\}$, and let
$N_\Omega=\sum_m|\mathcal X_m|$. The pooled full-field temperature error used
in the main paper is
\begin{equation}
E_{\mathrm{field}}
=\frac{\Delta T}{N_\Omega}
\sum_{m\in\{h,s,c\}}\sum_{\mathbf x\in\mathcal X_m}
\left|T_m^{\mathrm{pred}}(\mathbf x)-T_m^{\mathrm{ref}}(\mathbf x)\right|.
\label{eq:sup-mae}
\end{equation}
Because the temperature networks output dimensionless values, multiplication
by $\Delta T$ converts the error to kelvin. Equation~\eqref{eq:sup-mae}
matches the reported \texttt{volume\_mae}: every valid evaluation point
contributes once, rather than first averaging the three media.

For outlet quadrature points $\mathbf x_{f,k}$ with area weights $a_{f,k}$,
define a bulk-temperature estimator using a specified velocity field
$\mathbf v_f$:
\begin{equation}
\mathcal B_f[T;\mathbf v_f]
=
\frac{\sum_k a_{f,k}\rho_f
(\mathbf v_f(\mathbf x_{f,k})\!\cdot\mathbf n_f)
T(\mathbf x_{f,k})}
{\sum_k a_{f,k}\rho_f
(\mathbf v_f(\mathbf x_{f,k})\!\cdot\mathbf n_f)}.
\label{eq:sup-bulk-operator}
\end{equation}
The headline outlet metric is the mean absolute bulk outlet-temperature error
averaged over the hot and cold streams. It uses the same reference velocity
field in the prediction and reference estimators:
\begin{align}
e_{\mathrm{out},f}
&=\Delta T\left|
\mathcal B_f[T_f^{\mathrm{pred}};\mathbf u_f^{\mathrm{ref}}]
-\mathcal B_f[T_f^{\mathrm{ref}};\mathbf u_f^{\mathrm{ref}}]
\right|,
\label{eq:sup-outlet-stream}\\
E_{\mathrm{out}}
&=\frac12\left(e_{\mathrm{out},h}+e_{\mathrm{out},c}\right).
\label{eq:sup-outlet}
\end{align}
The common mass-flux weights isolate outlet temperature accuracy from
differences among independently trained flow networks. $E_{\mathrm{out}}$ is
reported in kelvin.

\subsection{Heat-Rate Accuracy and Device Closure}

For end-to-end energy diagnostics, each prediction instead uses its own
velocity field in Eq.~\eqref{eq:sup-bulk-operator}. Let
$\chi_f=C_f/C_{\min}$. The normalized hot-side heat release and cold-side heat
gain are
\begin{align}
Q_h
&=\chi_h\left(T^b_{h,\mathrm{in}}-T^b_{h,\mathrm{out}}\right),
\label{eq:sup-hot-rate}\\
Q_c
&=\chi_c\left(T^b_{c,\mathrm{out}}-T^b_{c,\mathrm{in}}\right).
\label{eq:sup-cold-rate}
\end{align}
where $T^b_{f,\mathrm{out}}
=\mathcal B_f[T_f^{\mathrm{pred}};\mathbf u_f^{\mathrm{pred}}]$ for a
prediction and the analogous reference fields are used for CFX. The
normalization is $Q_0=C_{\min}\Delta T$.

The reference-normalized two-sided heat-duty error is
\begin{equation}
E_Q
=100\sqrt{
\frac{
(Q_h^{\mathrm{pred}}-Q_h^{\mathrm{ref}})^2
+(Q_c^{\mathrm{pred}}-Q_c^{\mathrm{ref}})^2
}{2}
}\%.
\label{eq:sup-heat-rate-error}
\end{equation}
Unlike a closure residual, $E_Q$ evaluates whether the amount of heat
transferred on each side agrees with the CFX reference.

The independently evaluated device imbalance is
\begin{equation}
E_{\mathrm{clo}}=100|Q_h^{\mathrm{pred}}-Q_c^{\mathrm{pred}}|\%.
\label{eq:sup-ebal}
\end{equation}
This outlet-based metric is independent of the paired-wall closure term used
for training. None of $E_{\mathrm{out}}$, $E_Q$, or $E_{\mathrm{clo}}$ is
included in the training objective; all three are evaluated only after the
fixed training budget.

\section{Complete Results}
\label{app:results}

This section complements the main comparison with the complete TPMS heat-duty results.

\subsection{Reference-Normalized Heat-Duty Accuracy}

Table~\ref{tab:sup-heat-duty} reports $E_Q$ for every method--topology pair. Unlike device closure, this metric compares the predicted heat duty on both streams against the CFX reference and therefore measures whether the transferred energy is quantitatively accurate.

\begin{table}[t]
\centering

\scriptsize
\setlength{\tabcolsep}{4pt}
\begin{tabular}{lrrrr}
\toprule
\textbf{Method} & \textbf{P} & \textbf{G} & \textbf{D} & \textbf{IWP} \\
\midrule
E-MPINN & \underline{3.03} & 3.76 & 13.09 & 58.15 \\
RoPINN-CHT & 16.12 & 6.89 & 13.81 & 14.43 \\
CoPINN-CHT & 3.79 & 6.61 & 12.96 & 167.03 \\
MUSA-PINN-CHT & 7.95 & \underline{3.14} & \underline{1.63} & \underline{5.74} \\
\textbf{Cova-PINN} & \textbf{1.57} & \textbf{2.92} & \textbf{1.49} & \textbf{4.21} \\
\bottomrule
\end{tabular}
\caption{\textbf{Reference-normalized two-sided heat-duty error $E_Q$ (\%) on four TPMS exchangers.}
Lower is better; bold and underline denote the best and second-best results. Each value evaluates the complete end-to-end model produced by the corresponding training protocol.}
\label{tab:sup-heat-duty}
\end{table}

\subsection{Full Longitudinal Field Comparison}
\label{app:d_results}

Figure~\ref{fig:sup-d-longitudinal-all} extends the Diamond mid-plane comparison in the main paper to every baseline. Under matched temperature and absolute-error color ranges, the figure shows how prediction quality evolves through all five streamwise unit cells rather than only at the outlet.

\begin{figure*}[t]
\centering
\includegraphics[width=\textwidth]{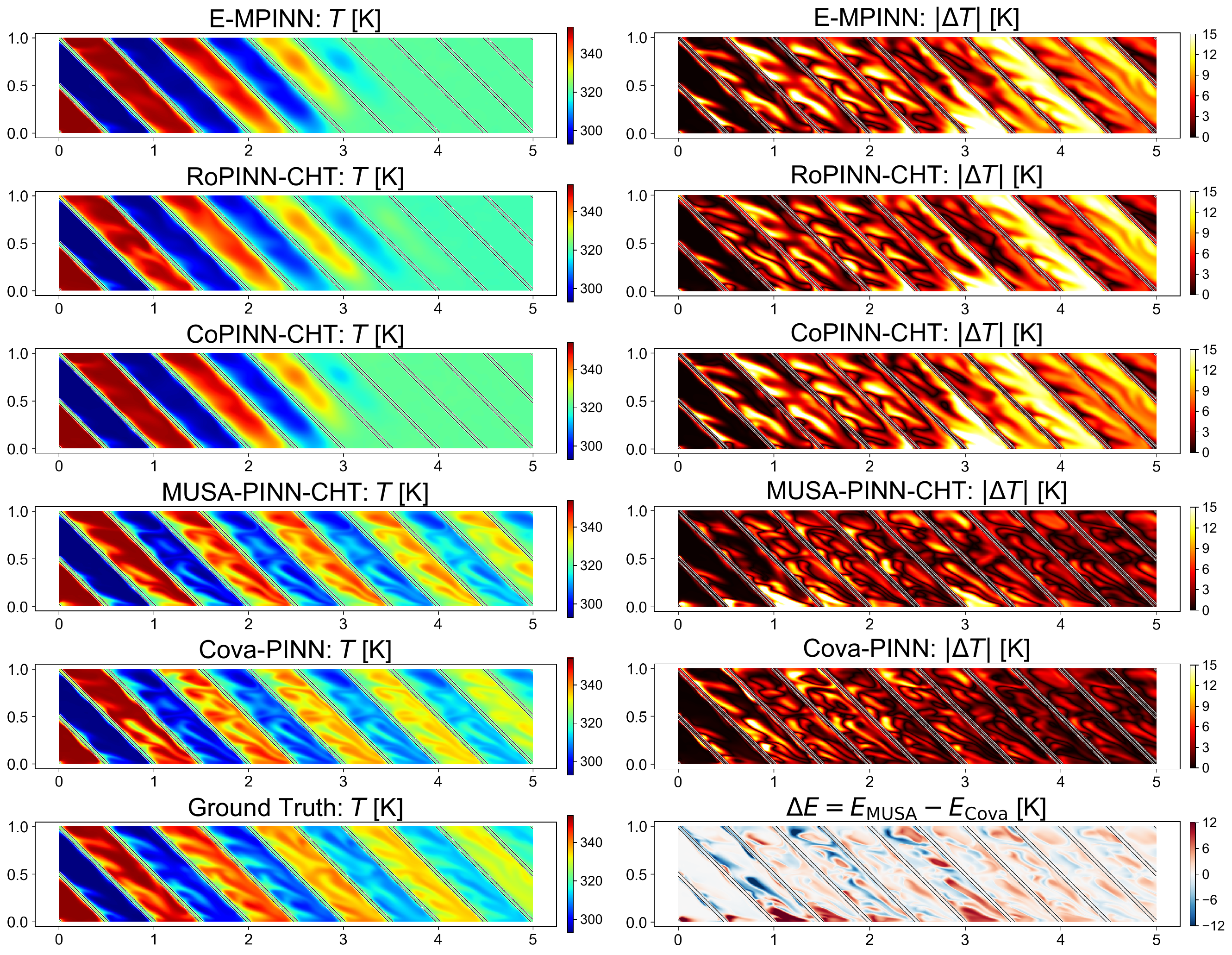}
\caption{\textbf{Full longitudinal comparison on the Diamond exchanger at $z=0.5$.}
The left column shows the predicted temperature fields of all methods and the CFX reference; the right column shows the corresponding pointwise absolute-error fields. Temperature and absolute-error panels use matched color ranges. The final panel reports
$\Delta E=|T_{\mathrm{MUSA}}-T_{\mathrm{CFX}}|-|T_{\mathrm{Cova}}-T_{\mathrm{CFX}}|$;
positive red values indicate lower Cova-PINN error, whereas negative blue values indicate lower MUSA-PINN-CHT error.}
\label{fig:sup-d-longitudinal-all}
\end{figure*}

\section{Computational Cost}
\label{app:cost}

We report wall-clock time to quantify the additional optimization cost of the staged Cova-PINN pipeline and its conservation quadrature. All measurements use one NVIDIA A40 GPU and the fixed $10{,}000$-update protocol described in Sec.~\ref{app:configuration}.

\begin{table}[t]
\centering

\scriptsize
\setlength{\tabcolsep}{4pt}
\resizebox{\columnwidth}{!}{%
\begin{tabular}{llrr}
\toprule
\textbf{Method} & \textbf{Training stage} & \textbf{Updates} & \textbf{Time (h)} \\
\midrule
E-MPINN & Joint & $10{,}000$ & $20.43$ \\
RoPINN-CHT & Joint & $10{,}000$ & $27.87$ \\
CoPINN-CHT & Joint & $10{,}000$ & $21.85$ \\
MUSA-PINN-CHT & Joint & $10{,}000$ & $32.76$ \\
\midrule
Cova-PINN (flow) & Flow stage & $5{,}000$ & $25.12$ \\
Cova-PINN (thermal) & Thermal stage & $5{,}000$ & $16.18$ \\
\textbf{Cova-PINN (total)} & \textbf{Sequential total} & \textbf{$10{,}000$} & \textbf{$41.30$} \\
\bottomrule
\end{tabular}}
\caption{\textbf{End-to-end wall-clock cost under a common $10{,}000$-update schedule.}
All runs use one NVIDIA A40 GPU. Baselines train jointly, whereas Cova-PINN reports its sequential flow and thermal stages together with their total.}
\label{tab:sup-cost}
\end{table}

Times measure the fixed training schedule rather than time to a convergence threshold. The comparison matches total update count, not wall-clock time or the number of updates applied to each individual field. Cova-PINN's additional cost reflects its staged optimization and method-specific local-control-volume and paired-wall quadrature.

\disableappendixtoc
\fi

\end{document}